\documentclass[review]{elsarticle}
\usepackage{amsmath}
\usepackage{helvet}  
\usepackage{courier}  
\usepackage{graphicx}
\usepackage{algorithm}
\usepackage{algorithmic}
\usepackage{subfigure}
\usepackage{xcolor, soul}
\sethlcolor{green}
\usepackage{lineno,hyperref}
\usepackage{float}
\usepackage{multirow}
\usepackage{threeparttable}
\modulolinenumbers[5]

\journal{}

\begin{document}

\begin{frontmatter}

\title{Can Language Representation Models Think in Bets?}

\author{Zhisheng Tang, Mayank Kejriwal}
\address{Information Sciences Institute \\ 
USC Viterbi School of Engineering\\
4676 Admiralty Way 1001, Marina Del Rey, California 90292}
\ead{\{zhisheng,kejriwal\}@isi.edu}




\begin{abstract}
In recent years, transformer-based language representation models (LRMs) have achieved state-of-the-art results on difficult natural language understanding problems, such as question answering and text summarization. As these models are integrated into real-world applications, evaluating their ability to make rational decisions is an important research agenda, with practical ramifications. This article investigates LRMs' rational decision-making ability through a carefully designed set of decision-making benchmarks and experiments. Inspired by classic work in cognitive science, we model the decision-making problem as a bet. We then investigate an LRM's ability to choose outcomes that have optimal, or at minimum, positive expected gain. Through a robust body of experiments on four established LRMs, we show that a model is only able to `think in bets' if it is first fine-tuned on bet questions with an identical structure. Modifying the bet question's structure, while still retaining its fundamental characteristics, decreases an LRM's performance by more than 25\%, on average, although absolute performance remains well above random. LRMs are also found to be more rational when selecting outcomes with non-negative expected gain, rather than optimal or strictly positive expected gain. Our results suggest that LRMs could potentially be applied to tasks that rely on cognitive decision-making skills, but that more research is necessary before they can robustly make rational decisions. 

\end{abstract}


\end{frontmatter}

\section{Introduction}

Transformer neural network-based \emph{language representation models (LRMs)}, such as the Bidirectional Encoder Representations from Transformers (BERT) \cite{devlin2018bert} and the Generative Pre-Trained Transformer (GPT) series of models \cite{radford2019language, brown2020language}, have led to impressive advances in natural language understanding. They have significantly advanced state-of-the-art performance on a variety of natural language tasks, ranging from information extraction  \cite{gupta2022matscibert} and semantic role labeling  \cite{larionov2019semantic}, to text summarization  \cite{aksenov2020abstractive}, cross-lingual and multi-lingual understanding \cite{lample2019cross}, and question answering  \cite{izacard2020leveraging}. Variants of such models  \cite{he2020deberta, li2020unimo, clark2020electra, khashabi2020unifiedqa} currently underlie the most successful systems on competition leaderboards hosted by the Allen Institute for Artificial Intelligence for several important benchmarks \cite{leaderboard}, \cite{wang2022modern}. Domain-specific versions of these models have also achieved impressive performance in their respective domains (e.g., scientific literature, patents and intellectual property, and biology). Representative examples include PatentBERT \cite{patentbert}, DistilBERT \cite{distilbert}, BioBERT \cite{lee2020biobert}, DocBERT \cite{docbert}, K-BERT \cite{kbert}, and SciBERT \cite{scibert}. More recently, they have also been applied in multi-modal settings involving both text, and visual modalities, such as video \cite{sun2019videobert}, \cite{lu2019vilbert}. 

Owing in part to the close connection between language and cognition \cite{beckage2016language,harris2006language}, a growing body of research is seeking to deduce the cognitive abilities (or lack thereof) of LRMs \cite{wallace2019nlp}, \cite{balasubramanian2020s}. There are both theoretical and practical reasons for this interest. The latter is important because these models are continuing to be integrated into, or otherwise used for, real-world applications and architectures in multiple enterprises and domains \cite{industry1,industry2,industry3,industry4}. The former is also important and may be attributed to the empirical success and rapid advancement of these models. Particularly, as these models continue to get larger, and are proving to be capable in ways that had not been initially conceived \cite{wei2022emergent}, there is rising interest in their fundamental properties, such as the dependence of their performance on size and number of parameters \cite{kaplan2020scaling}, their robustness \cite{wang2021measure}, including susceptibility to various flavors of adversarial attacks  \cite{fursov2022differentiable}, and the amount of knowledge `encoded' into their learned representations  \cite{heinzerling2020language}. This line of research is detailed further in Section \ref{rw}, where we discuss related work.         

Along these lines, it is unclear if LRMs can be trusted to make approximately rational decisions, even when the outcomes are  defined but are uncertain. In both behavioral and decision science, a number of experiments over the years have sought to test this ability in humans \cite{bscience1,bscience2,bscience3,sonsino2002rationality}. Classic work by Kahneman and Tversky showed, for instance, that people tend to exhibit loss aversion (prospect theory), and are not completely rational \cite{kahneman1979d}, \cite{kahneman2013prospect}. Despite this sensitivity to loss, however, prospect theory still predicts that an ordinary person would still choose gain over loss, if the gain substantially outweighs the loss in an equi-probable bet. Furthermore, people would not willingly choose to lose, if there was no possibility of gain attached to the choice. It is an open question whether LRMs take the same `commonsense' approach to making decisions, or (in contrast) they are prone to making decisions that would be considered extremely irrational in humans, such as actively pursuing loss (or to a lesser extent, zero gain) when a clear alternative is available with minimal risk.  

This article proposes a detailed set of research questions for empirically investigating the rational decision-making abilities of established transformer neural network-based LRMs. Inspired by longstanding work on behavioral science research mentioned above, the problem of making rational decisions is framed as one of testing a given model's ability to \emph{think in bets} and choose the outcome with the maximum expected gain. Specific contributions are enumerated below:

\begin{enumerate}
    \item We propose a novel set of research questions for understanding LRMs' rational decision-making abilities. To the best of our knowledge, this is the first such attempt to quantify LRMs' ability to make rational decisions by experimentally probing their capacity to `think in bets.'  
    \item We construct and present a novel set of decision-making and preference elicitation benchmarks for empirically investigating our research questions in a robust manner. The benchmarks are designed to actively preempt common issues with such models, such as superficial pattern matching (`shortcut learning') and dataset bias, which has also been found to be severe in other deep learning applications, such as computer vision \cite{du2022shortcut,ke3,tommasi2017deeper}.    
    \item We present a detailed and replicable methodology underlying the experimental study for investigating the presented research questions. Our methodology aims to control for several important factors that could serve as explanations for subsequently observed findings. Where applicable, we present new metrics and evaluation protocols to address such issues. At least one of these metrics, called Belief Conditioned Accuracy (BCA), is a novel metric that is specifically designed to quantify an LRM's rationality on a decision-making problem, after conditioning on a theoretically important predictor.
    \item We conduct an extensive experimental study investigating each of the proposed research questions. Each study is accompanied by a detailed set of statistics, and appropriate baselines. We conduct the study on four transformer-based LRMs that are established models already being incorporated in several industrial products and services. As noted earlier, we also emphasize robustness (through multiple evaluation metrics and statistics) in the experimental design itself. Using multiple evaluation metrics, we find that our core conclusions are largely consistent with one another. 
\end{enumerate}

The rest of this article is structured as follows. Section \ref{rw} discusses relevant related work, while Section \ref{sec:rq} enumerates the specific research questions that fall within the scope of this article. Section \ref{sec:materials} details the materials and methods underlying the study, and Section \ref{sec:results} follows with the key results and findings for each of the research questions in Section \ref{sec:rq}. Section \ref{sec:discussion} contextualizes these results with a broader discussion. Section \ref{sec:conclusion} provides some guidance on promising avenues for future research before concluding the article. 

\section{Related Work}\label{rw}

This article is primarily influenced by two broad bodies of research: fundamental research on transformer-based LRMs, and experimental studies investigating the properties of these LRMs. The latter area is especially relevant to our goals. We discuss each of these in turn below, with specific focus on work that is best related to our research objectives. Additionally, our experimental methodology, and the manner of benchmark construction, are influenced significantly by decision-making experiments in the behavioral sciences \cite{kahneman2013prospect,bscience1, sonsino2002rationality}. However, these experiments were mainly conducted on humans (and at times, in domain-specific settings such as science \cite{bscience2}). This article seeks to follow similar principles in benchmark construction, but applies the benchmarks on LRMs. Because the article considers LRMs as the primary objects of study, we begin by describing some key models that are also employed in the empirical study in this article, followed by detailing other recent work on understanding the properties of these models, and transformer-based models more generally.

\subsection{Transformer-based Language Representation Models (LRMs)}

As we noted in the previous section, language representation models (LRMs), which are also called neural language models, have achieved great success on a variety of natural language understanding tasks over the last half decade.   An early and influential LRM is Bidirectional Encoder Representations from Transformers (BERT), which uses a novel attention mechanism to obtain rich `pre-trained' representations of language from a large corpus of text. Pre-training is followed by task-specific `fine-tuning' that allows it to get state-of-the-art performance (at the time) on specific tasks, such as question answering, without requiring full re-training from scratch. In experiments, BERT was found to obtain a score of 80.5\% on the General Language Understanding Evaluation (GLUE) benchmark \cite{wang2018glue}, an improvement of 7.7\% over the previous best performing model. Similarly, it achieved an F1 score of 93.1\% on the Stanford Question Answering Dataset (SQuAD) v1.1 benchmark \cite{rajpurkar2016squad} and 83.1\% on SQuAD v2.0 \cite{rajpurkar2018know}, with improvements of 1.5\% and 5.1\% compared to the previous leading model, respectively. Moreover, the pre-trained BERT model was released publicly and is amenable to fine-tuning on other tasks. We draw on this capability in the proposed work. 

A more mature version of BERT is the Robustly Optimized BERT Pretraining Approach (RoBERTa) \cite{liu2019roberta}. RoBERTa is structurally the same as BERT. However, RoBERTa improves the training process on some key fronts, such as a bigger batch size, more extended sequence, and longer training. RoBERTa also removes the next sentence prediction objective and introduces the dynamic masking strategy. Therefore, compared to BERT on their published GLUE and SQuAD performance, RoBERTa shows significant improvements and obtains new state-of-the-art results on four of the GLUE tasks. Like BERT, the pre-trained RoBERTa model was released and is amenable to being fine-tuned. 

A more advanced version is the Decoding-enhanced BERT with disentangled attention (DeBERTa) \cite{he2020deberta}. DeBERTa is structurally similar to BERT and RoBERTa. DeBERTa also introduces several novel techniques to improve performance even further. First, DeBERTa uses a disentangled attention mechanism, where two vectors are used to represent the content and relative position of each word, and correspondingly, disentangled metrics are used to compute the attention weights. Second, during pre-training, DeBERTa uses an enhanced mask encoder to combine the absolute positions in the decoding layer to predict the masked tokens. These novel methods improve DeBERTa's performance on benchmark tasks. Compared to the \emph{RoBERTa-Large} model, DeBERTa improves on SQuAD v2.0 by 2.3\%, on Multi-Genre Natural Language Inference (MNLI) \cite{williams2017broad} by 0.9\%, and on Large-scale ReAding Comprehension Dataset From Examinations (RACE) \cite{lai2017race} by 3.6\%. The pre-trained DeBERTa is also publicly available and amenable to being fine-tuned. 

Another version is the Transformers for Longer Sequences (BigBird) model \cite{zaheer2020big}. BigBird deploy a sparse attention mechanism to reduce the quadratic dependency on sequence length, to linear dependency. The quadratic dependency on sequence length (in terms of memory) is one of the core limitations of transformer-based models, and necessary because of the full attention mechanism. As a result of its novel sparse attention mechanism, BigBird can handle up to 8x longer sequences using similar hardware, while demonstrating impressive improvements on question answering benchmarks.

Although the four models above are cited as representative examples, LRMs have continued to advance and become ever larger even in the last two years. Examples of recent LRMs include the Generative Pre-trained Transformer 3 (GPT-3) \cite{brown2020language}, the Language Models for Dialog Applications (LaMDA) \cite{thoppilan2022lamda}, and the Scaling Language Modeling with Pathways (PaLM) \cite{chowdhery2022palm}. However, due to their track record over the last five years, and their more manageable size, BERT-based models have tended to be incorporated in real-world systems, including within Google's own search engine \cite{BERTGoogle}. Therefore, for the experimental studies in this article, we use the four models detailed earlier (BERT, RoBERTa, DeBERTa, and BigBird). Our benchmarks and methodology are applicable to other models that are capable of natural-language question answering, but we leave an investigation of the bigger models for future research. 

\subsection{Understanding the Properties of LRMs through Experimental Studies}

Owing to the success of such LRMs, a recent line of work has emerged on understanding their properties using rigorous empirical methodologies, many inspired by research first conducted in the behavioral sciences. Prior work following BERT, for instance, has proposed approaches to better study the \emph{knowledge} encoded within these deep transformer-based LRMs. Examples include `fill-in-the-gap' probes for understanding the masked language model facility in LRMs \cite{rogers2020primer, wu2019mask}, probing of other classifiers that take different BERT representations as their feature-inputs \cite{liu2019linguistic, warstadt2020can}, deeper analysis of the self-attention weights in the LRMs \cite{kobayashi2020attention, ettinger2020bert}, and even a check-list style approach for comprehensively evaluating the linguistic abilities of a BERT-based model \cite{ribeiro2020beyond}. Detailed evidence suggests that BERT-based models seem to be encoding a `hierarchy' of linguistic features, with surface features at the bottom, syntactic features in the middle and semantic features at the top \cite{jawahar2019does}. Using massive corpora of pre-training text, the model is  able to learn such a hierarchy implicitly without requiring explicit training labels. 

While some work has found that information can be recovered from BERT's token representation \cite{wu2020perturbed}, the model still has trouble `understanding' concepts that are relatively natural to humans, such as negation and basic numeracy \cite{wallace2019nlp}. Like many other machine learning models, the model can also be overly confident in some of its inputs, and is susceptible to problems of both generalization and adversarial attacks \cite{fursov2022differentiable}, \cite{ke1}, \cite{ke2}, \cite{misra2022semantic}, \cite{ke3}.  Furthermore, several experiments have demonstrated that, although BERT effectively encodes information about relations, entity types, relations, semantic roles, as well as proto-roles, it can lose some of its robustness in the face of basic named entity replacements \cite{balasubramanian2020s}. 

This article contributes to this line of work by specifically investigating preference elicitation and rational decision-making abilities of such LRMs. We address the latter by posing bet questions to an LRM, and assessing its ability to think probabilistically, in terms of expected gains or losses, both when it is exposed to example bets (and allowed to `fine-tune' on them) and when it is not.  To the best of our knowledge, this is the first study to propose such an investigation, although there has been recent work on \emph{applying} LRMs to the sequential decision-making problem \cite{li2022pre}, which is reminiscent of planning (rather than behavioral decision-making, as studied herein). However, much more recently, language models have been playing a growing and prominent role of late in cognitive science research \cite{misra2021language,sawayama2022watching}. This work is intended to add to this growing body of research.

\section{Research Questions (RQs)}\label{sec:rq}

This article proposes to investigate three specific research questions (RQs), the first of which investigates the preferences of neural language models when provided with pairs of items that are of `high' or `low' value,  while the other two investigate their ability to make rational decisions under different assumptions and experimental conditions. 

\begin{enumerate}
    \item {\bf RQ1 (Preference Elicitation):} Can LRMs be trained to prefer a high-value item (e.g., a diamond) over a low-value item (e.g., a plastic pen), where value is understood in commonsense economic terms?  
    
    \item {\bf RQ2 (Thinking in Bets Without Task-Specific Fine-tuning):} Are LRMs able to rationally bet on outcomes with higher expected gain without first being fine-tuned on such bet questions?    
    
    \item {\bf RQ3 (Thinking in Bets With Task-Specific Fine-tuning):} Are LRMs able to rationally bet on outcomes with higher expected gain after being fine-tuned on such bet questions?
    
\end{enumerate}

While the first question is not (in itself) central to the goals of this work, it is an important prerequisite for investigating the other two questions. This is because the concept of rationality is linked to an agent's belief, or expectation, about value. For example, if an agent believes that an object A is more valuable than B, then given a bet with equi-probable outcomes (such as a coin toss), it is rational for the agent to `bet' on the outcome with A as the prize. In other words, the agent's preference of A over B influences our judgment of whether it subsequently makes a rational bet or not. Hence, understanding whether (and under what conditions) the language model's preferences align with our own is an important research question to address prior to investigating the model's ability to bet on rational outcomes. In exploring this question, we also compare the trained model's ability to that of a `default' LRM that has not been trained specifically for preference elicitation (but that still performs well on general commonsense question answering tasks) to quantify the effect of training itself on preference elicitation. 

The second question directly considers the issue of whether LRMs are able to think in bets without first being fine-tuned on bet questions. As discussed subsequently, we refer to this as `task-specific fine-tuning' wherein a model has been fine-tuned on a training set of questions that mirrors the purpose of the RQ, compared with the `default' version, as mentioned above. However, we also investigate  whether an LRM that has been fine-tuned for the preference elicitation task (RQ1) and performs well on it, is able to `naturally' think in bets, even without first being fine-tuned on bet questions.

The third question investigates whether the model is able to think more effectively in bets, both in absolute terms and relative to RQ2, once it undergoes such task-specific fine-tuning. Although not stated directly in the question itself, we also investigate the LRMs' generalization on bet questions that are \emph{structurally} similar but have different \emph{surface} form compared to questions that were used for task-specific fine-tuning. By structurally similar, we mean that the mathematical form of the bet, including the number of outcomes, and the probabilities associated with the outcomes, remains the same. An example would be two bets of the form where one involves tossing a coin, and the other of which involves randomly picking a card (from the standard 52-card deck). An outcome is then associated with whether the coin comes up heads or tails (for the former), or whether the randomly picked card is black or red (for the latter).  Assuming that the bet-wager and outcomes are identical in the two cases, the bets described above are structurally similar, but with different natural language descriptions associated with them. The third question attempts to quantify whether, and to what extent, the LRMs' decision-making ability erodes when the surface form of the question changes to one that it has not seen during fine-tuning. 

As noted in Section \ref{rw}, our research questions are heavily inspired by similar experiments in decision science and psychology, many of a classic nature \cite{kahneman1979d}; namely, we seek to understand a language model's rationality, and its ability to seek outcomes with maximum expected gain, by studying its preferences on a carefully designed set of prompts. It is important to design the prompts to control for a range of problems that are known to occur with such language models, including their sensitivity to format \cite{zhao2021calibrate}, and their propensity to achieve high performance through advanced, but ultimately superficial, statistical pattern matching \cite{du2022shortcut}. Hence, the experimental design, benchmark construction, and evaluation methodology are critical elements of the study, and are extensively detailed in the following section. 

\section{Materials and Methods}\label{sec:materials}

\subsection{Language Representation Models (LRMs)}

As discussed in Section \ref{rw}, many of the recent advances in language representation models (LRMs) are based on transformer  neural networks \cite{vaswani2017attention}. In some instances in the literature, these are referred to as language representation \emph{learning} models, or even \emph{neural} language models. We adopt the uniform terminology of language representation models in this article, with the understanding that we are primarily interested in the recent neural models. 

LRMs, such as BERT \cite{devlin2018bert} and the GPT \cite{radford2019language} series of models, have been found to generalize on an impressive range of language understanding tasks, including machine translation and question answering \cite{joshi2017triviaqa, zellers2018swag, yang2018hotpotqa, tiedemann2016finding}. In the remainder of this article, we uniformly use the term LRM to refer to the models that are used to answer `multiple choice' prompts or questions by selecting one answer from a set of candidate choices.

To be applied to specific NLP problems, these models, which are \emph{pre-trained} on a large corpus of text before they are publicly released, are typically also \emph{fine-tuned} on an additional smaller data set to optimize them for the task at hand. For example, if BERT were to be applied to the problem of Named Entity Recognition or NER (automatically extracting named entities, such as people, places and organizations, from text), the pre-trained version would have to be fine-tuned on a `training' set of clearly defined NER inputs and outputs. Fine-tuning takes a much smaller amount of time compared to  pre-training. This makes pre-trained LRMs a powerful asset in the NLP literature because they can be used as a `base' model for a wide range of tasks and data sets, a facility that we rely upon for the decision-making experiments herein.

Owing to its training on a large body of text, the pre-trained model can be fine-tuned to `score' a natural language sentence based on its likelihood of being a plausibly constructed sentence. The higher the score, the more plausible the sentence. Impressively, the score correlates not just with real-world syntactic usage but also plausible semantics, depending on the background corpus on which the model was pre-trained. For instance, if the model was pre-trained on a general corpus, such as Wikipedia or Google Books, nonsensical sentences would tend to be given much lower scores by the model. However, in some cases, the pre-training corpus is domain-specific, such as with the BioBERT pre-trained model \cite{lee2020biobert} or social media-based pre-training \cite{Zafarani+Liu:2009}. Such models will normatively assign higher scores to biological and social media sentences, respectively. We only use models in this article that were pre-trained on general corpora, such as Wikipedia and news articles. 

Although there are many viable transformer models (and their variants) available at the time of writing, we selected four models for our studies, first introduced and discussed in Section \ref{rw}: BERT \cite{devlin2018bert}, RoBERTa  \cite{liu2019roberta}, DeBERTa \cite{he2020deberta} and BigBird \cite{zaheer2020big}. We emphasize again that RoBERTa is fundamentally similar to BERT, but is often treated separately because of its (much) higher performance over the original BERT release owing to its robust optimization, and other important engineering innovations \cite{liu2019roberta}. Our rationale for selecting these four LRMs is that they are established models that have been rolled out in a range of commercial and outward-facing products, including the Google search engine \cite{BERTGoogle} and Amazon Web Services \cite{AWSBERT}. Many technical and domain-specific variants of these models have also been developed and deployed, including SciBERT \cite{beltagy2019scibert}, BioBERT \cite{lee2020biobert} and AlBERTa \cite{lan2019albert}. 


An important commonality between these four models is that they all have pre-trained versions publicly available, but can also be fine-tuned on additional question answering data sets. Since the training data set used for fine-tuning depends on the experiment and research hypothesis, we specify the data set used for fine-tuning when discussing the experimental methodology for the corresponding research hypotheses. Next, we describe the specific manner in which each of these LRMs can be applied to the Multiple Choice Question Answering (MCQA) problem, which is of central interest in this article.

\subsection{Multiple-Choice Question Answering (MCQA) using LRMs}
\label{mcqa-ft}

In this section, we introduce some basic formalism on MCQA instances, and on the specific methodology that we use to obtain an LRM's prediction for a given instance. An MCQA instance formally consists of two elements: a `question' prompt $q$ and a set $C$ of $n$ `answer' choices $\{c_1, c_2, ..., c_n\}$. We assume, without loss of generality, that each of the choices and $q$ is represented as a string. Furthermore, it is usually assumed that exactly one of the choices in $C$ is `correct'. Given a set of MCQA instances (referred to as an MCQA \emph{benchmark}), the goal of a question-answering system, such as an appropriately fine-tuned LRM, is to predict the correct choice for the prompt. 

One approach by which an LRM can be made to answer an MCQA instance is as follows. First, the question prompt $q$ is concatenated with \emph{each} of the choices $c_i$ in turn. This yields $n$ question-answer pairs, where a pair $p_i = concatenate(q, c_i)$. Next, each of the $n$ pairs is fed into the model in turn (i.e., independent of one another) during the fine-tuning phase, when the correct answer can be revealed to the model. Specifically, if the $c_i$ used to form $p_i$ is the correct choice, $p_i$ is labeled as 1 (otherwise it is labeled as 0). Given such a `training' set, the model is fine-tuned to minimize the cross-entropy loss as is standard for MCQA problems.

The fine-tuned model can then be evaluated using a similar methodology on unseen MCQA instances, for which it needs to predict the correct answers.  
First, we convert such a `test' MCQA instance to a similar input structure, as used during fine-tuning, by concatenating each choice $c_i$ to the question prompt $q$ (to obtain pair $p_i$). Next, the model is provided with each $p_i$ \emph{independently}, and outputs a score for each such pair. The score is assumed to be proportional to the model's belief in that pair being labeled as 1 in the underlying ground-truth. Because the model's score is not necessarily normalized, we use the sigmoid function\footnote{The sigmoid function is $f(x)=\frac{1}{1+e^{-x}}$. Here $x$ is the `raw' score output by the final layer of the model. The output of the sigmoid function is guaranteed to lie in $[0,1]$ and is monotonic in $x$.}, to normalize each score to the range [0,1]. Although just the highest-scoring choice could be selected as the model's prediction, there is an alternative mechanism available for selecting (and evaluating) the predictions in a decision-making context, as later discussed. For this reason, it is more appropriate to say that, depending on the specific experiment and research question, a \emph{predicting function} is applied to the $n$ normalized scores (corresponding to the $n$ choices in an MCQA instance) to yield the model's prediction for that instance. 

As a concrete example, consider the MCQA instance in Table \ref{table:mcqa}. The fine-tuned LRM would be given each question-answer pair in turn. If the predicting function is simply to select the choice for which the model outputs the highest score, and this choice happens to be `This statement is true: airplane is more expensive than pen', then the model would (correctly) select the choice `airplane is more expensive than pen' as its prediction, given the prompt `This statement is true:'.

\begin{table}[H]
\resizebox{\columnwidth}{!}{%
\begin{tabular}{ c c c }
    Prompt & Choice & Question-Answer Pair \\
    \hline
    \multirow{3}{*}{This statement is true:} & airplane is more expensive than pen & This statement is true: airplane is more expensive than pen \\\cline{2-3}
    & pen is more expensive than airplane & This statement is true: pen is more expensive than airplane \\\cline{2-3}
    & airplane and pen have the same value & This statement is true: airplane and pen have the same value \\\cline{2-3}
    \hline
\end{tabular}%
}
\caption{An example of a Multiple-Choice Question Answering (MCQA) instance (prompt and choice-set) and the concatenated pairing of the choices in turn that is used to evaluate the LRMs in this paper.}
\label{table:mcqa}
\end{table}

We place no constraints at present on the predicting function: it may select zero, one, or more than one, choice, as the prediction for a MCQA instance. In the next section, we describe two plausible choices for the function, one of which is to just select the highest-scoring choice.   


An alternative approach to fine-tuning a model for question answering is to concatenate the question prompt $q$ with \emph{all} of the choices $(c_1, c_2, ..., c_n)$ together, along with a separator between each of the choices i.e., $(1, 2, ..., n)$. This yields a single `complete' multiple choice question, which is denoted as $ mcq = concatenate(q, 1, c_1, 2, c_2, ..., n, c_n)$. Instead of 1 or 0, the expected output (or label) of the model should be the \emph{string} of the correct choice, denoted as $c_T$. The actual fine-tuning process is similar in that the model generates string output given the $mcq$, and the cross-entropy loss between $c_T$ and this string output is used in the optimization. 
However, while this can work well for `generative' question answering problems, models such as BERT are better suited for discriminative problems (of which the MCQA problem is one) and typically adopt the first approach. While language models, such as UnifiedQA, have also been applied to generative QA problems \cite{khashabi2020unifiedqa}, we leave a generative evaluation of a model's decision-making ability for future work and assume the first or `discriminative' approach in the rest of the paper.

\subsubsection{Predicting Functions: Standard Method and Threshold Method}

An obvious choice for the predicting function that we had mentioned briefly earlier, applicable when exactly one `correct' prediction is desired from the model, is to select the choice $c_i$ corresponding to the highest normalized score. We call this function the \emph{standard method}, as it is the method favored in much of the QA literature where exactly one choice is correct and all other choices are incorrect. An example presenting the method in action was earlier presented in the context of Table \ref{table:mcqa}.

However, in the decision-making benchmarks that are considered in this article, the assumptions about correctness are more nuanced. For example, while our benchmark construction (subsequently described in Section \ref{sec:BC}) always guarantees that there is a single `optimal' answer, it is not always the case that all other answers are equally sub-optimal. Some choices provided with a bet question may be associated with positive expected gain (even though they may not be associated with the \emph{highest} positive expected gain, which would be the case for the optimal answer) while others may be associated with zero, or even negative, expected gain. An important empirical objective in this paper is to determine whether the model is able to understand these differences, especially when it selects sub-optimal choices. 

Hence, to more holistically evaluate the model's decision-making ability, we also consider a second predicting function called the \emph{threshold method}. As the name suggests, instead of simply selecting the choice with the highest score, this method involves selecting all choices (as predictions) that lie \emph{above} a threshold.  Furthermore, if no choice lies above the threshold, the model refuses to yield a prediction. In fact, given $n$ choices in an MCQA instance, it is easy to show that a model can theoretically yield the power-set (of size $2^n$) of selections of choices, of which the empty set is one extreme possibility, and the complete set (selecting all choices) is the other extreme possibility. 

In principle, this methodology is similar to that employed in a wide range of practical machine learning applications that require the careful selection of a threshold in order to optimize a non-trivial quality metric (such as F1-score) on a multi-label problem \cite{Fan2007ASO}. Similarly, when describing the benchmark construction in Section \ref{sec:BC}, we systematically consider how to evaluate the quality of a model when using this method, but for now, it suffices to say that there is more than one reasonable way to construct a `ground truth' against which to evaluate an LRM's (multi-label) predictions when using the threshold method.  We define such a `binary' ground-truth as stating which of the $2^n$ possible power-set predictions should be considered as correct or incorrect, on the basis of which an accuracy metric can always be computed for a model. For example, one choice of ground-truth might consider as correct any combination of choices (that a model selects, using the threshold method) that yields positive expected gain, while another (less conservative) ground-truth may only decide to test for consistency (i.e., it may only penalize a set of selected choices that are directly contradictory, such as `bet on heads' and `do not bet').  The rationale and specific rules governing  ground-truth construction for a corresponding benchmark will be provided in Section \ref{sec:BC}.

Because there is more than one way to judge the quality of such multi-label predictions, an `optimal' choice for this threshold is not a fixed value, and depends not only on the manner (including the choice of ground-truth) in which the model's performance is being judged, but also on the benchmark itself. Furthermore, a threshold that works well for one model (and for a given quality metric) may be  sub-optimal under a different experimental condition. For all of these reasons,  instances in each of our benchmarks are always partitioned into a train, development and test set. Where applicable, the train set is used for fine-tuning, and without exception, the test set is always used for evaluating all models under a given experimental condition to ensure fair comparison. Similarly, when using the threshold method as a predicting function, the development set is always used to determine the `optimal' threshold, given an LRM, and choice of ground-truth against which the LRM's predictions will be judged.  

To discover such a threshold, which is technically a \emph{hyperparameter} that takes values in [0,1], we first do a simple grid search in that range using increments of 0.01. In the event that more than one threshold value achieves the maximum performance on the development set, we select the median of the values that achieve this maximum as the expected optimal threshold for that experiment.

\subsubsection{Fine-Tuning LRMs for MCQA}
\label{fine-tuned qa lrms}

As described earlier, the pre-trained versions of the LRMs need to be fine-tuned on appropriate `training' sets before they can be applied on tasks with a specific structure. MCQA is an example of such a task (and is the primary focus of this article) but other common examples in the NLP literature include named entity recognition \cite{liu2019knowledge} and information extraction \cite{gupta2022matscibert}. The four pre-trained LRMs that we fine-tuned for the MCQA experiments in this paper are BERT, RoBERTa, DeBERTa and BigBird, all of which were introduced earlier. Note that these LRMs have variants in the HuggingFace repository that we used for accessing and fine-tuning the models. The specific variants that we used are $BERT_{BASE}$ \cite{hfbert}, $RoBERTa_{BASE}$ \cite{hfroberta}, $DeBERTa_{BASE}$ \cite{hfdeberta}, and $BigBird_{BASE}$ \cite{hfbigbird}.

Although we could directly fine-tune each of these pre-trained models on the train set of the MCQA benchmark that we construct, one of our empirical goals is to understand whether such models, if fine-tuned on a `general purpose' MCQA benchmark, are able to exhibit reasonable  decision-making ability as a natural consequence of such fine-tuning. One such benchmark that is widely used in the community is the Situations With Adversarial Generations (SWAG) data set \cite{zellers2018swag}. SWAG is a commonsense benchmark that contains MCQA instances on grounded commonsense inference and physically grounded reasoning. 

By fine-tuning each of the four LRMs on SWAG, we can test whether good performance on such commonsense tasks necessarily entail good performance on decision-making and preference elicitation problems expressed using everyday language and objects. Since these SWAG-based fine-tuned models form a natural basis of comparison to models that are further fine-tuned to handle our decision-making benchmarks, we refer to them as the \emph{default} models. For example, the \emph{default BERT} model is used to refer to the BERT model that has been fine-tuned on the training partition of the SWAG benchmark. 

For the fine-tuning itself, we use a batch size of eight and fine-tune each of the four models for three epochs (and a total of 27,500 steps) each on the 73,546 MCQA instances in the SWAG training set, using a learning rate of 5e-5. Following the fine-tuning, we verified that, on the SWAG validation set, the accuracy of the default BERT, RoBERTa, DeBERTa and BigBird model is 77\%, 79\%, 85\% and 81\%, respectively. These results are consistent with previously published results and confirm that the models are indeed able to achieve good performance after fine-tuning. Finally, we uniformly use a batch size of 32 and a learning rate of 5e-5 for any other (i.e., non-SWAG) fine-tuning, described below, in our experiments. 

Because the default model can be \emph{further} fine-tuned on another MCQA benchmark with a similar structure, we use it as the `initialization' for fine-tuning an LRM on the train set of our own benchmark (which depends on the specific research question). We mnemonically refer to such a model as a \emph{task-specific fine-tuned LRM}. Since each of our benchmarks is always partitioned into train, development and test sets, we use the train set for the actual fine-tuning, and we stop the fine-tuning once the model has achieved an accuracy of 90\% on the development set. A key point that we emphasize here is that, unlike the default model (of which there is only one unique model per LRM, since it is never fine-tuned on a decision-making benchmark), the task-specific fine-tuned model depends  on the benchmark that was used for further fine-tuning. Hence, there is a unique task-specific fine-tuned model per LRM \emph{and} benchmark. Finally, when evaluating a (default or task-specific fine-tuned) model using the threshold method, the development set is `re-used' for determining the expected optimal threshold for that model using the grid-search procedure described earlier.

\subsection{Benchmark Construction}\label{sec:BC}

\subsubsection{High-Value and Low-Value Sets of Items}\label{bc-items}

We manually create two sets of `high-value' and `low-value' items to facilitate our experiments. These items are tabulated in Table \ref{table:itemsets}. We created these sets with the intent that any regular person would be able to distinguish these items fairly easily, especially if asked to do so in terms of (difference in value in) dollar amounts. We recognize that `value' can be understood in different ways, and can even be contextually dependent e.g., a `low-value' item can always become high-value in the right set of circumstances, and vice versa. We adopt a commonsense, everyday view here, with value best thought of in economic terms. This is a standard premise in the decision-making literature, going back to the pioneering behavioral psychology experiments devised by Kahneman and Tversky \cite{tversky1974judgment}. Also, it is the \emph{distinction} (or relative difference) between high and low value items that we control for in our experiments, not the absolute value of these items. Furthermore, in our experimental design, we take into account the potential concern that an LRM may not understand `value' in quite the way described above. We present questions using several different \emph{templates}, including making the economic aspect of value explicit in one of these templates. This construction is discussed further in the next section.

\begin{table}[H]
\begin{tabular}{ p{3cm} p{8cm} }
    \bf Train Set \\
    \hline
    High value items & airport, airship, bike, bicycle, bus, camera, gold, supercar, refrigerator, jewelry, hotel, horse, guitar, tank \\
    \hline
    Low value items & baseball, bread, brush, chair, chocolate, vegetable, soup, shirt, orange, knife, fish, cookie, cigarette, honey, newspaper \\
    \hline
    \\ 
    \bf Development Set \\
    \hline
    High value items & watch, ipad, phone, tv, telescope \\
    \hline
    Low value items & egg, apple, soda, toothbrush, toothpaste \\
    \hline
    \\
    \bf Test Set \\
    \hline
    High value items & car, house, diamond, airplane, computer \\
    \hline
    Low value items & pen, paper, water, slipper, sock \\
    \hline
\end{tabular}
\caption{High-value items and low-value items, divided into train, development and test sets.}
\label{table:itemsets}
\end{table}

The high-value items are partitioned into `train', `development', and `test' high-value sets (and similarly for the low-value items). 
The roles of the train and development sets were noted earlier in Section \ref{mcqa-ft}. Unless explicitly noted, the outcome of the experiment will be reported using the test set. Importantly, this is true even when we are not fine-tuning the model on our train set i.e., when we are using a default model. 
This allows us not only to use \emph{paired} statistical significance testing when applicable, but to use common item-sets as the basis for all experiments. This ensures that any differences observed between experiments cannot be explained through differences in benchmarks used during testing.

\subsubsection{Value Questions}\label{bc-vq}
Given the high-value items and low-value items described in the previous section, RQ1 seeks to determine whether an LRM prefers (at least, on average) a high-value item over a low-value item. One manner in which we can do this is by prompting an LRM to choose the more `valuable' item from a given pair of items (one of which could be high-value and the other of which could be low-value). More precisely, we could design a \emph{template}, defined as a partial instance with placeholders that is converted into an actual instance by being appropriately \emph{instantiated}. An example of a template would be the partial instance: `This statement is true: (a) [h] is more expensive than [l] (b) [l] is more expensive than [h] (c) [h] and [l] have the same value' Here, [h] and [l] are placeholders for a high-value and low-value item from Table \ref{table:itemsets}, respectively. To differentiate between a template and its instantiated equivalent, we refer to an instantiated template as a \emph{value question}.

To reasonably control for format and potential different interpretations of `value', we designed four different templates for determining LRMs' preferences.  Controlling for format is motivated by the fact that LRMs are known to be sensitive to the mode of presentation of a prompt and answer choices, as evidenced in previous work \cite{jin2020bert}. Using only one template runs the risk of drawing conclusions about a model, even though the model itself may be quite capable of distinguishing between differently valued items if the inputs were presented in a slightly different format. At the same time, it is obviously not feasible to try \emph{every} possible format for investigating whether a model truly prefers a higher-value item `h' to a lower-value item `l'. Hence, we decided on a pragmatic compromise by constructing four templates that are structurally dissimilar; yet, consistently likely to yield results that would allow us to confidently distinguish between LRMs' preferences.

\begin{table}[H]
\begin{tabular}{ p{1.6cm} | p{4.5cm} | p{4.5cm}}
    \multicolumn{3}{c}{\bf} \\
    \bf  Name & \bf Template & \bf Example Instantiation (Value Question) \\
    \hline
    Boolean Expensive & This statement is true: (a) [h] is more expensive than [l] (b) [l] is more expensive than [h] (c) [h] and [l] have the same value & This statement is true: (a) car is more expensive than pen (b) pen is more expensive than car (c) car and pen have the same value \\
    \hline
    Boolean Valuable & This statement is true: (a) [h] is more valuable than [l] (b) [l] is more valuable than [h] (c) [h] and [l] have the same value & This statement is true: (a) car is more valuable than pen (b) pen is more valuable than car (c) car and pen have the same value \\
    \hline
    Choice Expensive & From [h] and [l], choose an item that is more expensive: (a) [h] (b) [l] (c) the same & From car and pen, choose an item that is more expensive: (a) car (b) pen (c) the same \\
    \hline
    Choice Valuable & From [h] and [l], choose an item that is more valuable: (a) [h] (b) [l] (c) the same & From car and pen, choose an item that is more valuable: (a) car (b) pen (c) the same \\
    \hline
\end{tabular}
\caption{The four templates, mnemonically named, along with an example instantiation (value question) that respectively substitutes an actual high-value item and low-value item for [h] and [l] in the corresponding template.}
\label{table:value-question}
\end{table}

These four templates are listed in Table \ref{table:value-question}, along with an example instantiation using a high-value and low-value item combination from Table \ref{table:itemsets}. While the most important difference between the templates is the manner in which answer choices are presented (\emph{Boolean} versus \emph{Choice}), another important difference is that two templates use the word `valuable', while two others use the word `expensive' (\emph{Valuable} versus \emph{Expensive}). We anticipate that using the four templates derived from these two formatting controls will allow us to gain a more robust understanding, and conduct consistency checks, of the differences between LRMs' preferences for pairs of (high-value and low-value) items sampled from Table \ref{table:itemsets}.

Note that a value question associated with a high-value item `h' and a low-value item `l' can have different ground-truths depending on the choice of predicting function assumed:
\begin{enumerate}
    \item {\bf Standard method:} For each value question, the only correct choice in the ground-truth is the one implying that (depending on the template)  `h' is more \emph{valuable} or \emph{expensive} than `l'. All other choices are incorrect.
    
    \item {\bf Threshold method:} Recall that, per this method, the model may select zero, one or multiple (including all) choices as its prediction. Such a multi-label prediction can always be represented as a \emph{set} of choices. An appropriate,  binary ground-truth must operate at the level of sets, rather than choices. Since each value question always has three choices, there are $2^3=8$ possible predictions. We design three different ground-truths to determine which of these eight predictions should be considered as `correct' or `incorrect', each of which successively relies on a `broader' definition of correctness.

    \begin{enumerate}
      \item \emph{Normal:} Analogous to the ground-truth used for the standard method, a model's prediction is deemed as correct if it returns the set that contains only the one correct answer per the standard method (i.e., depending on the template, that `h' is more valuable or expensive than `l'). Note that accuracy, measured using this ground-truth, can theoretically be different from the accuracy measured using the standard method. One reason (although not the only one) is that the maximum-scoring answer may have a score that is below the (empirically determined) threshold, in which case no answer is selected and the empty set is returned.
      
      \item \emph{Weak Normal:} This ground-truth takes a broader view of correctness than the Normal ground-truth. In addition to what the Normal ground-truth deems correct, this ground-truth also rates the model's prediction as correct if it selects both the correct answer (per the Normal ground-truth) \emph{and} the answer that states that the two items are equal in value. We interpret this output as the model predicting that the high-value item (in the question) $\geq$ the low-value item. Any other combination of selected answers (as well as the empty set) beyond the two possibilities stated above would be treated as incorrect. 
      
      \item \emph{Weak:} This ground-truth takes the broadest view of model correctness. Unlike the other two ground-truths, it does not test whether the model is outputting the correct value preference; rather, it tests whether the model is either contradicting itself or otherwise refusing to state a preference. Hence, only the following three predictions are considered as incorrect (and each of the other five possible sets are considered as correct): (i) the set containing all three choices, (ii) the empty set, and (iii) the set containing the two `strict inequality' choices (expressing simultaneously that the value of `h' is strictly higher \emph{and} lower than the value of `l'). Intuitively, (i) and (iii) imply a `contradiction' while (ii) implies a non-decision that we treat as incorrect.
      
    \end{enumerate}
    
    Note that, because the three ground-truths successively rely on weaker notions of what counts as correct, the expected performance of a system that randomly selects from the eight possible sets correspondingly increases. Specifically, across each of the test sets used for evaluating RQ1 (regardless of template), expected random performance using the Normal, Weak Normal and Weak ground-truth is 12.5\%, 25\%, and 62.5\%, respectively. Expected random performance using the standard method is 33\%, which is (expectedly) different from the Normal ground-truth random performance, for the reason discussed earlier.
\end{enumerate}

\subsubsection{Bet Questions}
\label{bq}

While RQ1 investigates LRMs' preference for high-value items over low-value items, RQ2 aims to go a step further by evaluating whether LRMs are able to use this knowledge, whether directly or indirectly, to think in bets. Similarly, RQ3 aims to evaluate whether the models, if fine-tuned on bet questions, are able to successfully think in bets that they have not seen before. 

Both RQ2 and RQ3 require presenting to the model a bet question, with different \emph{uncertain} outcomes (each of which is associated with an expected gains), to investigate the degree to which the model is able to make a rational bet. Although both research questions evaluate the model on bet questions presented during testing, the key difference between them, discussed subsequently, is that RQ2 does not use bet questions for fine-tuning, while RQ3 does. 

We design the bet questions using a similar template-based methodology as the value questions. These templates contain placeholders for high- and low-value items (that can be instantiated using the items in Table \ref{table:itemsets}), but to expand and diversify the manner in which bets are presented to the model, we constructed questions using different bet \emph{modalities}, mnemonically designated as \emph{Coin}, \emph{Card} and \emph{Dice}, and demonstrated in Table \ref{table:bet-question}. Importantly, all three modalities involve equi-probable outcomes and are expressed using natural (rather than mathematical) language. We refer to an instantiated bet template as a \emph{bet question}.

\begin{table}[H]
\begin{tabular}{ p{3cm} p{8cm} }
    \bf Modality & \bf Question Template \\ 
    \hline
    Coin & If the coin comes up heads, then I win a [h]/[l]. If it comes up tails, then I lose a [l]/[h]. What should I do to maximize my expected gains? (a) I should bet on heads (b) I should bet on tails (c) I should not bet on either one \\
    \hline
    Dice & If the dice comes up even, then I win a [h]/[l]. If it comes up odd, then I lose a [l]/[h]. What should I do to maximize my expected gains? (a) I should bet on even (b) I should bet on odd (c) I should not bet on either one \\
    \hline
    Card & If I pick a card from a standard deck of cards, and the card is red then I win a [h]/[l]. If it is black, then I lose a [l]/[h]. What should I do to maximize my expected gains? (a) I should bet on red (b) I should bet on black (c) I should not bet on either one \\
    \hline
\end{tabular}
\caption{Templates expressing the three bet modalities used for evaluating both RQ2 and RQ3, that can be instantiated using a high- and low-value item (represented as [h] and [l], respectively). Note that a question cannot be instantiated using two high-value or two low-value items (i.e., if a high-value item is selected for the first placeholder, a low-value item must be selected for the second placeholder).}
\label{table:bet-question}
\end{table}

These bet questions are based on some reasonable assumptions that are expected to hold in practice and are consistent across all questions. First, even though it is not directly mentioned in the question, any model that chooses to bet is assumed to wager some amount of money on the outcome it chooses. For example, in the case of Coin questions, the model may choose to wager on either heads or tails (if it chooses to bet). Second, once the bet is executed and the outcome is known, the following standard reward structure is assumed: whatever outcome comes to pass, the win/loss event associated with the outcome (e.g., winning a watch if the coin comes up heads) will be executed, unless the model chose not to bet. Note, however, that in our experiments, we do not actually `simulate' the execution of the bet, since we are only interested in expected values. Regardless of the executed event, however, if the model bet on the outcome that came to pass, it will receive its wagered amount back, otherwise it will lose its wagered amount. 

Although we could have included the wagered amount explicitly, or made it a free parameter, doing so would have added the confounding element that the model, were it to perform badly, was doing so because of its limited ability to handle \emph{explicit numeracy}. Studies have shown that numerical reasoning can be problematic for some language models, and that it may need special treatment \cite{hendrycks2021measuring, lin2020birds}. To control for this, we rely on a construction that assumes (approximately) \emph{balanced expectation} i.e., the wager is an amount that lies roughly between the high- and low-value items, but is skewed slightly toward the lower end to allow for some outcomes to have positive expected gain associated with them. More technically, if we denote the value of a prototypical high- and low-value item as H and L respectively, the wagered amount X is assumed to obey the inequality $L <= X <= 0.5*(H-L)$. 

With the above assumption in place, we calculate expected gain for the different bet questions (reproduced in the Appendix), with ground-truths constructed accordingly. In all cases, this construction guarantees that there is always (exactly) one optimal-expectation outcome. However, depending on the question, there may be outcomes that have positive expected gain, but are not necessarily optimal. Furthermore, because we allow the model to choose not to bet, the optimal-expectation outcome is never negative; however, in some cases, not betting is optimal, with zero gain, since all other outcomes are associated with negative expected gain. As discussed in the next section, we evaluate the model under several different scenarios and metrics to gain a more comprehensive and robust understanding of its decision making, rather than always expecting it to choose the one outcome that is strictly optimal.

Finally, we instantiate and design the templates in a way that minimizes the possibility of model-overfitting due to (potential) superficial pattern matching. One way in which we do so is by flipping the position of `win' and `lose' in the templates shown in Table \ref{table:bet-question} to avoid a fixed outcome always being the optimal (or even positive expected gain) choice. Formally, given $m$ and $n$ high- and low-value test items, respectively, the total number of Coin-modality test questions would be $2 \times 2 \times m \times n = 4mn$, and similarly for the other modalities. While the first doubling effect accounts for the win/lose swapping (not shown in the template), the second doubling effect accounts for the fact that the first [h/l] placeholder could have either a low-value or high-value item. 

In some cases, not betting is clearly the rational choice. For example, considering the Coin question template, if the model wins a low-value item and loses a high-value item, then under the assumption about the wager stated earlier, not betting is rational. In the next section, we formalize this intuition by presenting RQ-specific metrics that quantify LRMs' ability to distinguish between these possibilities.

Similar to the value questions, the ground-truth for the bet questions depends on the choice of predicting function used during evaluation:
\begin{enumerate}
    \item {\bf Standard method:} Because each choice is associated with a expected gain, the only correct choice is the one that maximizes expected gain. Each of the other two choices is deemed as incorrect. Note that the construction of the benchmark guarantees that there is always one unique choice with maximum expected gain (which is always non-negative).
    
    \item {\bf Threshold method:} Because the model now returns a set of choices (leading to $2^3=8$ possible predictions), each such prediction needs to be systematically assigned an expected gain. Note that, when the model returns the set consists of the two `I should bet' choices (expressing that I should bet on both outcomes e.g., heads \emph{and} tails) as prediction, we assume that the wagered amount is split equally between the selected choices, and we calculate the expected gain of such prediction accordingly. For all other prediction possibilities, we provide a detailed calculation of the expected gain in the Appendix. While exactly one choice (per bet question) is always associated with an optimal gain, which has the highest expected gain among all the possible predictions, the other choices are not necessarily equally sub-optimal. To investigate the LRMs' decision making abilities holistically, we design three ground-truths to determine the correctness of the eight multi-label predictions.
    
    \begin{enumerate}
      \item \emph{Strict:} Similar to the `Normal' ground-truth in the benchmark used for RQ1, the `Strict' ground-truth only deems a model to have correctly answered a bet question if it returns the set that contains only the optimal choice (i.e., the choice with the highest expected gain). If the optimal choice is not in the set, or choices besides the optimal choice are in the set, the model's prediction is deemed as incorrect. For reasons similar to those stated earlier, the accuracy measured using the `Strict' ground-truth can differ from the accuracy measured using the standard method.
      
      \item \emph{Positive Gain:} This ground-truth measures the model's ability to return a set, such that the choices in the set collectively yield a positive expected gain, when such a set exists. Calculations for which such predictions would have positive expected gain, under the wager assumption stated earlier, are provided in the Appendix. Note that, for some bet questions, each of the eight possible predictions is associated with either zero or negative expected gain. Since there is no `correct' prediction possible for such questions (when evaluating a model using this ground-truth), we exclude such bet questions when evaluating a model using this ground-truth. Furthermore, any set containing choices that are either collectively contradictory, as well as the empty set, is deemed as incorrect. Hence, the following predictions would always be considered incorrect: (i) the set containing all three choices, (ii) the empty set, and (iii) the set containing the choice `I should not bet on either one' \emph{and} any one of the other two choices. 
      
      \item \emph{Non-Negative Gain:} This ground-truth measures the model's ability to return a set containing choices that collectively yield a non-negative expected gain (including zero expected gain). This ground-truth is the broadest of the three ground-truths discussed thus far. However, unlike the Positive Gain ground-truth, all questions have at least one correct set associated with them, since the choice of not betting is always associated with zero expected gain. Note also that the three scenarios listed for the Positive Gain ground-truth, whereby a prediction would always be considered incorrect if it implied non-decision or contradiction, also apply to this ground-truth.
    \end{enumerate}

\end{enumerate}

Similar to the RQ1 benchmark, because the three ground-truths successively take a broader view of what predictions count as correct, the expected performance of a system that randomly selects from the eight sets, correspondingly increases. Specifically, the expected random performance using the Strict, Positive Gain and Non-Negative Gain ground-truths is 12.5\%, 25\% and 25\%, respectively. Expected random performance using the standard method is still 33\%.

\subsection{Experimental Setup}
\subsubsection{RQ1: Preference Elicitation}
\label{vq}

We use the value-questions benchmark described in Section \ref{bc-vq} for investigating RQ1. The full benchmark comprises four different templates, each of which is instantiated using the train, development and test items described in Section \ref{bc-items}. We report results for the default, and task-specific fine-tuned models, using both the \emph{standard method} and the \emph{threshold method} introduced in Section \ref{mcqa-ft}. We emphasize that, while there is a single default model per LRM, there is a task-specific fine-tuned model per LRM \emph{and} per template. Hence, the full RQ1 evaluation involves a total of four default LRMs, and 16 task-specific fine-tuned LRMs.

We use the accuracy metric for reporting performance. A single accuracy estimate is reported for the standard method for each experimental setting (i.e., choice of LRM and template). In contrast, three accuracy estimates are reported for the threshold method, with each estimate corresponding to each of three ground-truths (Normal, Weak Normal and Weak). In a slight abuse of terminology, we use the name of the ground-truth itself to refer to the corresponding accuracy of the model being evaluated using that ground-truth.

We also report on the statistical significance of each result by using the one-sided z-test to evaluate whether the result is better than the expected random performance i.e., the expected accuracy of a system that selects randomly from among the answer choices. For the standard method, the expected random performance is 33\% for all experimental settings in RQ1, since only one out of three possible choices can be selected by any system being evaluated using the standard method (and exactly one choice is correct). For the threshold method, we similarly computed the fraction of correct answers for each of the three ground-truths, which would equal the expected random performance. We found the expected random performance for the Normal, Weak Normal and Weak ground-truth to be 12.5\%, 25\%, and 62.5\%, respectively. In the results, we use a maximum Type-I error rate $\alpha = 0.05$ to confirm significance (in other words, the one-sided P value, using the z-test, must not exceed 0.05 to be significant); however, the complete set of P values for all tests are also reproduced in the Appendix.

\subsubsection{RQ2: Thinking in Bets Without Task-Specific Fine-Tuning}
\label{bq-1}

We use the bet-questions benchmark described in Section \ref{bq} for investigating RQ2. The four default models used for RQ1 are also used for RQ2. Furthermore, since the (task-specific) fine-tuned models used for RQ1 behave similarly across the four templates (as the results for RQ1 will show), we only report results for the (RQ1) fine-tuned LRM that was fine-tuned on the \emph{Choice Valuable} template. In total, this yields four  fine-tuned models that are used for investigating RQ2. 

Note that, because of the nature of this research question, the three templates (each corresponding to a different modality i.e., Coin, Dice and Card; see Table \ref{table:bet-question}) contained in the benchmark only need to be instantiated for the development and test set items in Table \ref{table:value-question}. The former is only necessary for determining the optimal threshold for each evaluated model (four default and four fine-tuned) when using the threshold method, and is not needed when using the standard method.

The remainder of the testing procedure is similar to RQ1. When using the standard method, we report accuracy for each of the eight models described above, for each of the three modalities. Similarly, when using the threshold method, we report the three accuracy estimates (for each of the eight models) corresponding to the \emph{Strict}, \emph{Positive Gain}, and \emph{Non-Negative Gain} ground-truth described earlier.  

In addition to using the ordinary accuracy (which measures predictions against a ground-truth), this research question also seeks to assess whether a model's decision on a bet question can be explained by the model's value preference. For example, consider a model that has expressed the preference that a pen is more valuable than an airplane. While this preference would be treated as `incorrect' by the ground-truth, it could serve as the rational basis for the model's decision on a bet question using those items. Put simply, if the model \emph{believes} that a pen is more valuable than an airplane, then it would be \emph{rational} for the model to select `I should bet on heads' given the bet question `If the coin comes up heads, then I win a pen. If it comes up tails, then I lose a airplane. What should I do to maximize my expected gains?' 

As the example above demonstrates, since the ordinary accuracy (on the bet questions) is independent of the model's value preference or `belief', it may not be a fair estimate of the model's rationality. A better estimate can be obtained instead by measuring the model's accuracy (on bet questions) conditioned on its belief (on corresponding value questions). We refer to this metric as the \emph{belief conditioned accuracy (BCA)}, and report it using the standard method. The main difference between the BCA and the ordinary accuracy is that the former directly tests the model's rationality by treating the decision as `correct' if it is rational to do so, given the model's belief. 

Statistical significance results are reported using the same methodology as for RQ1 i.e., by comparing each result, using the one-sided z-test, to the corresponding expected random performance. For the standard method, the expected random performance is still 33\%, while for the Strict, Positive Gain and Non-Negative Gain ground-truth, it is 12.5\%, 25\% and 25\%, respectively.

\subsubsection{RQ3: Thinking in Bets After Task-Specific Fine-Tuning}\label{bq-2}

We use the same bet-questions benchmark as used in RQ2 for investigating RQ3. Recall that this benchmark contains bet questions in three different modalities (Coin, Dice, and Card). However, unlike RQ2, where the benchmark was only used for testing and development, the benchmark (appropriately instantiated) is used for training, testing and development for investigating RQ3.

Because the four default LRMs are already investigated in RQ2, we focus on the task-specific fine-tuned LRMs in RQ3. Specifically, we fine-tuned all four pre-trained LRMs on each of the three modalities, yielding a total of $4*3=12$ task-specific fine-tuned LRMs.

One of the main goals in RQ3 is to evaluate whether a task-specific fine-tuned LRM that is fine-tuned using one modality (e.g., Card) is able to generalize reasonably well to the other two modalities (e.g., Coin and Dice). Hence, each of the 12 (task-specific fine-tuned) LRMs is evaluated separately on each of the Coin, Card and Dice data sets instantiated using the test items.  This enables us to contrast an LRM's results when it is fine-tuned and tested on the same modality, versus a different modality. 

We report results for both the standard and threshold method. Similar to the previous RQs, the development set is used for determining when to stop fine-tuning the model, and for selecting appropriate thresholds. A single accuracy estimate is reported when using the standard method for each experimental setting. Three accuracy estimates are reported for the threshold method, each corresponding to a ground-truth (Strict, Positive Gain and Non-Negative Gain). Following RQ2, we use the name of the ground-truth itself to refer to the corresponding accuracy of the LRM when evaluated using that ground-truth.

Statistical significance results are reported using the same methodology as for RQ2 i.e., by comparing each result, using the one-sided z-test, to the corresponding expected random performance. The expected random performance for both the standard method and the threshold method (per ground-truth) are identical to those reported earlier in RQ2, since the test sets used in RQ2 are identical to those used in RQ3.

\section{Results}\label{sec:results}

\subsection{RQ1: Preference Elicitation}\label{result-rq1}

\begin{table}
\centering
\caption{Model accuracy (as percentage) for instantiated value questions using the four templates introduced in Table \ref{table:value-question}: Boolean Expensive (BE), Boolean Valuable (BV), Choice Expensive (CE), and Choice Valuable (CV). }
    \begin{threeparttable}
        \begin{tabular}{ ccccc }
             \hline
             Template $\rightarrow$ &BE & BV & CE& CV\\
             $\downarrow$ Model &\multicolumn{4}{c}{Default/Fine-tuned} \\
             \hline
             BERT  &\textbf{68}$^\dag$/96$^\dag$ &36/96$^\dag$ &\textbf{56}$^\dag$/\textbf{100}$^\dag$ &\textbf{52}$^\dag$/\textbf{100}$^\dag$ \\
             RoBERTa  &24/92$^\dag$ &48/\textbf{100}$^\dag$ &44/100$^\dag$ &44/100$^\dag$ \\
             DeBERTa  &4/\textbf{100}$^\dag$ & 8/96$^\dag$ &40/100$^\dag$ &48/100$^\dag$ \\
             BigBird  &0/100$^\dag$ & 4/100$^\dag$ &52$^\dag$/100$^\dag$ &48/100$^\dag$ \\
             \hline
        \end{tabular}
    
        \begin{tablenotes}\footnotesize
            \item[1] $\dag$ indicates that the result is statistically better than random performance (33\%) with 95\% confidence.
            \item[2] \textbf{Bold} text indicates the best result, if statistically significant, for the given column. When there are multiple perfect scores within a column, only the first result is in bold.  We report exact P values in the Appendix.
            \item[3] Where task-specific fine tuning is involved, the model is fine-tuned using the same template as for the evaluation.
        \end{tablenotes}
    \end{threeparttable}
    \label{table:rq1-result}
\end{table}

Recall that the first research question (RQ1) involves evaluating whether, or which of, the different LRMs are able to successfully distinguish between a `high-value' and `low-value' item that they have not encountered during fine-tuning. As described in Section \ref{vq}, four different templates are instantiated as `value questions' for the purposes of investigating RQ1 to ensure that the results are robust to different choices of question and answer format with (effectively) the same content. We report the results in Table \ref{table:rq1-result} for both the default and fine-tuned models. The results illustrate that, in the general case (i.e., for 12 out of 16 cases), the default models cannot distinguish between high-value and low-value items in a way that is (statistically) any better than random selection. However, there are some interesting exceptions. For instance, the performance achieved by the BERT default model (68\%, 56\% and 52\%) is significantly better than random for the Boolean Expensive, the Choice Expensive, and the Choice Valuable template, respectively. These results underscore the methodological decision to use four templates for robustly investigating RQ1, which must be (generally) borne in mind when evaluating LRMs because of their sensitivity to the particular form of input.

However, after they have been fine-tuned, the LRMs outperformed their default counterparts, and in all cases, achieve over 90+\% accuracy with statistical significance. In other words, these models can distinguish between a `high-value' and `low-value' item after they have been fine-tuned on a data set that follows the same template, but is instantiated using different item sets (Table \ref{table:itemsets}). This suggests that LRMs may already have the ability to distinguish between differently valued items, but need to be fine-tuned using appropriate prompts in order for us to access this ability (i.e., using similarly structured prompts, but with potentially different pairs of items). Additionally, if fine-tuning is permitted, the model also loses its sensitivity to the actual template. As shown in the table, performance is largely similar across the four templates for all the LRMs. 

\begin{figure}[H]
  \centering
  \includegraphics[width=\textwidth]{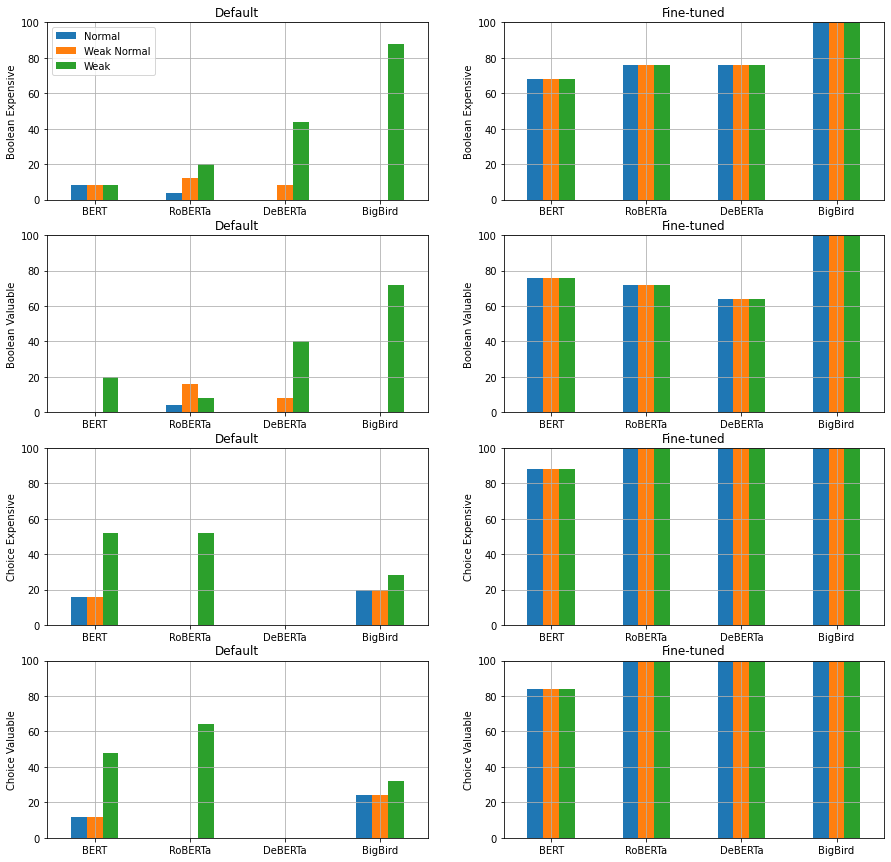}
\caption{The default and fine-tuned LRMs' accuracy using the threshold method (introduced in Section \ref{mcqa-ft}) and its associated ground-truths: Normal, Weak Normal, and Weak (introduced in Section \ref{bc-vq}). Questions are instantiated using the four templates introduced in Table \ref{table:value-question}: Boolean Expensive, Boolean Valuable, Choice Expensive, and Choice Valuable. Where fine tuning is involved, the model is always fine-tuned using the same template as used during testing.}
\label{figure:rq1}
\end{figure}

Figure \ref{figure:rq1} reports the accuracy using the threshold method and its associated ground-truths: Normal, Weak Normal, and Weak. Recall that the expected random performance for each of these ground-truths is 12.5\%, 25\%, and 62.5\%, respectively. Focusing on the default LRMs' results, we observe that they are unable to distinguish between high and low-value items better than random, regardless of which template and ground-truth is used, with few exceptions. In contrast, the fine-tuned LRMs exhibit performance that are well above random, and in some cases, near-perfect. Consistent with the previous result, this result suggests that LRMs may already have the ability to distinguish between high-value and low-value items, and that they only need to be fine-tuned using the appropriate prompt to access this ability. Interestingly, we also find fine-tuned LRMs' performance to be identical across the three ground-truths, although the absolute performance depends on the specific LRM, with the fine-tuned BigBird consistently achieving the highest performance.

\subsection{RQ2: Thinking in Bets Without Task-Specific Fine-Tuning}
\label{rq2-result}

RQ2 was designed to test whether LRMs have (at least approximately rational) decision-making ability when bet questions are used as prompts. We investigate whether LRMs that have \emph{not} been fine-tuned on bet questions, but are fine-tuned on value questions (also used for fine-tuning in RQ1), have such abilities. Note that the LRMs we used for investigating RQ2 are a subset of the LRMs used in RQ1. The default LRMS remain the same, whereas, for the fine-tuned LRMs, we only use the models fine-tuned on the `Choice Valuable' template, owing to the homogeneous performance of each fine-tuned model across the four templates. The key difference between RQ1 and RQ2 lies in the prompt that is input to the models during testing. While value questions (instantiated using test set items) were used for investigating RQ1, bet questions are used for RQ2.

\begin{table}[H]
\caption{LRMs' performance (expressed as percentage) for bet questions, defined in Table \ref{table:bet-question}, using ordinary accuracy (ACC) and belief conditioned accuracy (BCA), which was introduced in Section \ref{bq-1}.}
\begin{threeparttable}
\resizebox{\columnwidth}{!}{%
\begin{tabular}{ ccccccccc }
 \hline
 Modality $\rightarrow$ & \multicolumn{2}{c}{Coin} && \multicolumn{2}{c}{Dice} && \multicolumn{2}{c}{Card} \\
 \hline
 Metric $\rightarrow$ & ACC & BCA & & ACC & BCA && ACC & BCA \\ 
 $\downarrow$ Model& \multicolumn{8}{c}{Default/Fine-tuned} \\
 \hline
 BERT & 43$^\S$/\textbf{50}$^\S$ & 42$^\S$/\textbf{50}$^\S$ && 25/\textbf{52}$^\S$ & 25/\textbf{52}$^\S$ && 42$^\S$/\textbf{50}$^\S$ & 27/\textbf{50}$^\S$\\
 RoBERTa & 25/25 & 25/25 && 47$^\S$/25 & 43$^\S$/25 && 29/32 & 28/32\\
 DeBERTa & \textbf{49}$^\S$/44$^\S$ & \textbf{53}$^\S$/44$^\S$ && \textbf{50}$^\S$/50$^\S$ & 45$^\S$/50$^\S$ && \textbf{48}$^\S$/47$^\S$ & \textbf{47}$^\S$/47$^\S$\\
 BigBird & 35/25 & 39/25 && 48$^\S$/29 & \textbf{46}$^\S$/29 && 37/48$^\S$ & 35/48$^\S$\\
 \hline
\end{tabular}%
}
\begin{tablenotes}\footnotesize
\item Where fine-tuning is involved, the LRMs are fine-tuned on the (Choice Valuable) value questions introduced in Table \ref{table:value-question}. 
\item $\S$ indicates that the result is statistically better than random performance (33\%) with 95\% confidence. The complete set of P values is reproduced in the Appendix. 
\item \textbf{Bold} text indicates the best result, if statistically significant, for the given column and metric (e.g., ACC Default). 
\end{tablenotes}
\end{threeparttable}
\label{table:rq2-result-raw}
\end{table}

As discussed in Section \ref{bq}, three bet modalities are instantiated for investigating the research question more robustly. We report both the default and fine-tuned LRMs' performance in Table \ref{table:rq2-result-raw} using two metrics: ordinary accuracy (ACC) and belief conditioned accuracy (BCA). 

Considering first the default LRMs' performance using the ordinary accuracy metric, we find that the LRMs are not able to `correctly' answer bet questions. This result is qualitatively consistent with the results obtained for RQ1. However, on occasion, a better-than-random result is still obtained, such as for the default DeBERTa model on the \emph{Coin} questions. In general, default models' accuracy (whether ordinary or belief-conditioned) does not exceed 50\%. 

Fine-tuned accuracy can be higher for some models and modalities, but a clear trend is not distinguishable, and results are not always significant. For example, the BERT fine-tuned model achieves a significant 52\% performance on the \emph{Dice} modality (using both the ACC and BCA metrics) but the DeBERTa fine-tuned model achieves lower performance than even its default counterpart on the \emph{Coin} modality. Hence, unlike for RQ1, fine-tuning does not yield definitive performance improvements on the task. Even when such improvements are observed, they are relatively lower than the near-perfect results obtained for RQ1 following fine-tuning.

It is also interesting to note that the BCA performance, on average, does not (significantly and consistently) exceed ACC performance for any given model. Recall that the BCA is computed by constructing a `ground truth' that is based on the model's own beliefs. In the Appendix, we also report performance when evaluating the models using the threshold method, but the conclusions were found to be largely consistent with Table \ref{table:rq2-result-raw}. Namely, we found that default or (value-questions) fine-tuned LRMs are unable to correctly choose, on average, the option that maximizes the expected gain, or that even achieves positive expected gain.

One potential reason that a given model is unable to answer bet questions, despite being able to distinguish between high- and low-value items, may be its inability to understand the prompt at a syntactic level, since it has not been fine-tuned on bet questions. In RQ3, we investigate this hypothesis in more detail by evaluating models that have been fine-tuned on bet questions.

\subsection{RQ3: Thinking in Bets After Task-Specific Fine-Tuning}\label{ftlrms}

In contrast with the previous experiment, in this experiment, we evaluate the LRMs after fine-tuning them on bet questions. There are two experimental goals: first, does fine-tuning on bet questions, instantiated with one set of high- and low-value items (`train' set), improve performance on bet questions instantiated with a \emph{different} set of high- and low-value items (`test' set), similar to what was observed for RQ1? Second, how does the performance get impacted when fine-tuning is conducted using one modality (e.g., Card) but tested using a different modality (e.g., Dice)?

\begin{figure}[H]
  \centering
  \includegraphics[width=\textwidth]{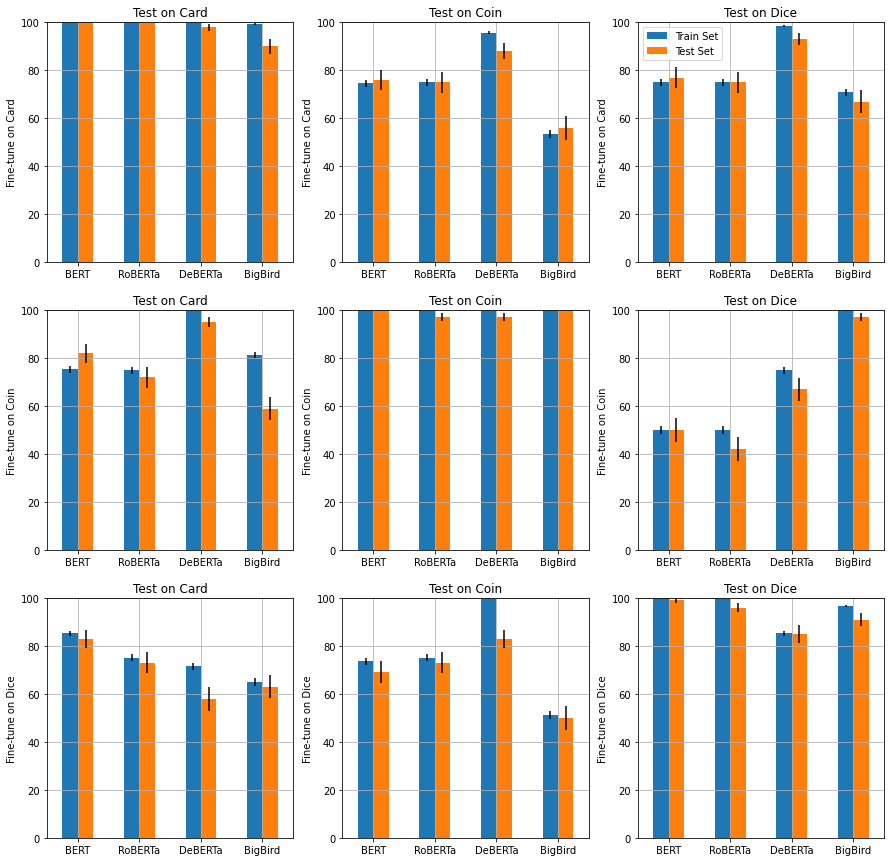}
\caption{The accuracy, with standard error bars (when non-zero), of fine-tuned LRMs on three instantiated bet-modality datasets (Card, Coin, Dice). Exact values are reproduced in the Appendix). The `train' and `test' sets are populated using the (respectively named) item sets in Table \ref{table:itemsets}. The train set is always used for fine-tuning, and in some experimental conditions (blue bars) is also used for testing. The modality used during fine-tuning is indicated along the vertical axis, while the modality used during testing is indicated along the horizontal axis. All results are statistically better than random performance (33\%) with 95\% confidence. The complete set of P values is reproduced in the Appendix.}
\label{table:rq2-result-fig}
\end{figure}

Figure \ref{table:rq2-result-fig} illustrates the accuracy for all four base models ($BERT_{BASE}$, $RoBERTa_{BASE}$, $DeBERTa_{BASE}$ and $BigBird_{BASE}$, introduced in Section \ref{fine-tuned qa lrms}), fine-tuned using the three different modalities on the `train' set items, and evaluated using the three modalities on both the `train' and `test' set items. Our reason for also evaluating each fine-tuned model on the train set is to assess the impact of modality on performance while controlling for the item-sets. We find that all results are statistically significant compared to random performance. Note that, since the test-set BCA was found to be exactly the same as the corresponding test-set (ordinary) accuracy, BCA results are not shown in the figure.

The diagonal plots in Figure \ref{table:rq2-result-fig} show that all fine-tuned LRMs achieve near-perfect performance regardless of whether the `train' or `test' set is used for evaluation. This provides some support for our earlier claim that there may be a strong dependency on the prompt and its format, and that the choice of item-set matters much less. In other words, when the bet modality is known in advance and can be used for fine-tuning, the performance of the model is expected to be high (at least for the equi-probable bet modalities considered here). Additionally, when we consider the figure as a whole and compare the `train' and `test' performance (blue versus orange) in each experimental setting, we find that there is no strong dependence on the choice of the item-set used during evaluation. Surprisingly, the model does not gain a noticeable advantage from `re-observing' the `train' item set during evaluation. This may be because we explicitly designed the benchmark to avoid potential overfitting: the same pair of items is used in several bet questions, but with different `optimal' outcomes. For example, in one bet question, a high-value item may be on the losing side of the bet, and a low-value item on the winning side (and vice versa, in another bet question). Therefore, the fine-tuned LRMs are unable to `memorize' their way to the correct answer, as would ordinarily be expected when the train-set is re-used during testing.

Turning to the two off-diagonal or \emph{cross-modal} entries in each row of the figure, we find that all four LRMs show about 25\% decrease in performance, compared to their respective diagonal entries. This provides further evidence of the dependence of performance on modality, regardless of item-set used. However, despite this relative decline in performance, the fine-tuned LRMs still achieve an average accuracy of around 70\% in most experimental settings, which is well above random performance and qualitatively within reach of state-of-the-art performance on many QA benchmarks \cite{rajpurkar2016squad}, \cite{yang2018hotpotqa}, \cite{joshi2017triviaqa}. The performance is especially striking compared to the results in the previous section, where most LRMs' performance could not be statistically distinguished from random performance, and the best performance was only 53\%. 

Additionally, there is no one LRM that was found to consistently out-perform the others when comparing models in each cross-modal experiment (or off-diagonal plot). For example, while DeBERTa exhibits the best performance in some cross-modal settings (e.g., when fine-tuning on Card and testing on Coin and Dice), BigBird exhibits near-perfect performance in the setting when the LRMs are fine-tuned on Coin and tested on Dice, and BERT shows the best performance when the LRMs are fine-tuned on Dice and tested on Card.

\begin{figure}[H]
  \centering
  \includegraphics[width=\textwidth]{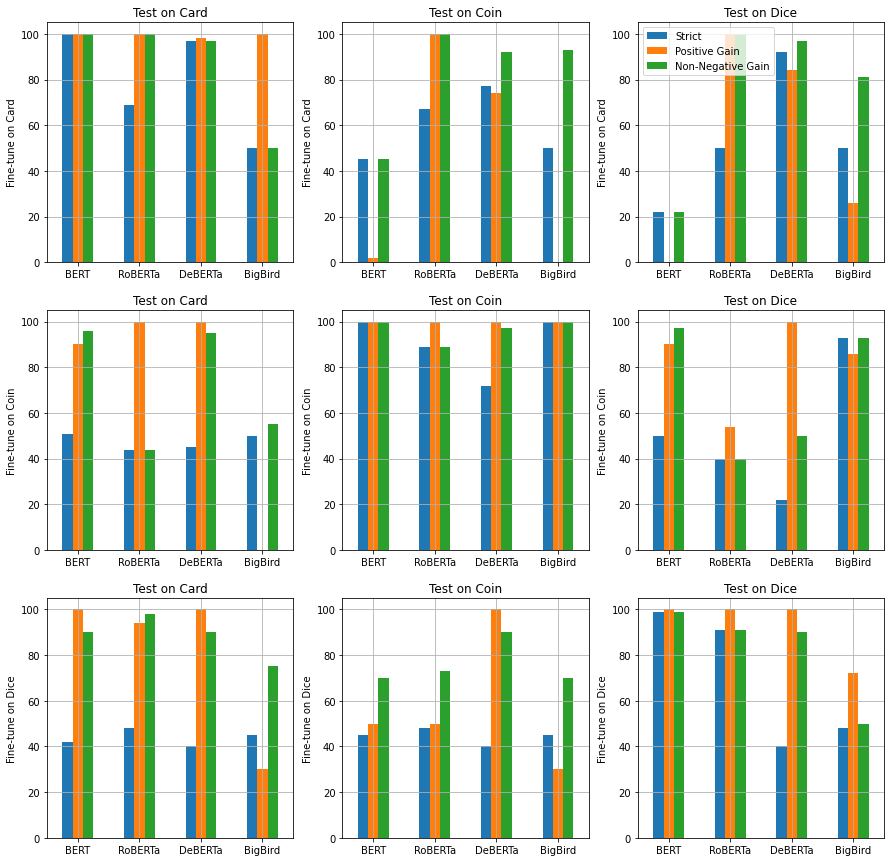}
\caption{The accuracy, using the threshold method, of fine-tuned LRMs on three instantiated bet-modality datasets (Card, Coin, Dice). During testing, the bet questions are always instantiated using the test set items in Table \ref{table:itemsets}, with the train set items always used for fine-tuning. The fine-tuning and evaluation modalities are indicated along the vertical and horizontal axis, respectively. The three ground-truths were introduced in Section \ref{bq}.}
\label{table:rq3-threshold}
\end{figure}

Similar to Figure \ref{table:rq2-result-fig}, Figure \ref{table:rq3-threshold} reports the accuracy using the threshold method using three different ground-truths: \emph{Strict, Positive Gain}, and \emph{Non-Negative Gain}. Results are reported for all modalities, but we do not include the results of evaluating the models using the `train' set, since the previous result established that the performance is largely similar (owing to the models' inability to over-fit to the `train' item-set). 

As earlier noted, because there are bet questions where the best answer only yields non-negative gain (since such questions have no positive expected gain answers), we exclude such questions when calculating accuracy using the Positive Gain ground-truth. Because of this filtering, it is theoretically possible for an LRM to have higher performance when evaluated against the Positive Gain ground-truth compared to the Non-Negative Gain ground-truth.

Similar to the previous result, the highest performance observed (across LRMs) in the diagonal entries is near-perfect, regardless of the ground-truth used. However, some models do much worse than others in the same experimental setting. For instance, we observe performance (using the Strict ground-truth) as low as 40\% for DeBERTa, when it is fine-tuned and evaluated on the Dice modality. 

In the cross-modal (off-diagonal) setting, we observe an average decline of more than 40\% when evaluating the LRMs using the Strict ground-truth. Compared to Figure \ref{table:rq2-result-fig}, this result suggests that the threshold method proves to be a more challenging evaluation paradigm for the LRMs than the standard method, at least when the optimal answer is expected (captured by the Strict ground-truth). As expected, the performance rebounds when the model is only expected to choose outcomes that have positive expected gain or non-negative expected gain. In line with the previous results, we find again that performance of the LRMs is strongly dependent on the modality used during fine-tuning, and that their ability to generalize to other modalities is limited. This claim is also indirectly supported by the considerable variance observed across models and settings, even when using the less conservative (Positive Gain and Non-Negative Gain) ground-truth. 

Focusing on the Positive Gain results in the cross-modal setting, we observe several near-zero entries (e.g., when BERT was fine-tuned on the Card modality and tested on the Coin and Dice modalities, as well as when BigBird was fine-tuned on the Coin modality and tested on the Card modality), because of the LRMs' inability to identify any outcome with positive expected gain, even when such outcomes are present among the candidate answer choices.

Interestingly, when evaluated using the Non-Negative Gain ground-truth, the LRMs do show some evidence of generalization. For instance, even in the cross-modal setting, we observe that their performance decreases (on the Non-Negative Gain ground-truth) by a lower margin than on the other two ground-truths. This result is consistent with the one in RQ2, suggesting that they are better (and also generalize better) at identifying non-negative outcomes rather than strictly positive and optimal outcomes.

\subsection{Summary of Results}

This section summarizes the key results from the previous section:
\begin{enumerate}
    \item In investigating RQ1 (Section \ref{vq}), we found that the LRMs can distinguish between low- and high-value items (with accuracy above 90\%), if they are fine-tuned on a train set with the same template as the test set, even if the latter is instantiated with items not seen before during fine-tuning. This result also suggests that our choice of high- and low-value items is not arbitrary, and that the model largely agrees with our distinction between these two item-sets. Without the template-specific fine-tuning, however, the default LRMs' performance was found to be statistically indistinguishable from random performance. 
    \item In investigating RQ2 (Section \ref{bq-1}), the results show that the default LRMs cannot make `rational' bets (with few exceptions) any better than random guessing. Fine-tuning on the value questions was not found to achieve any noticeable difference in performance. Even when evaluating LRMs' performance using Belief Conditioned Accuracy (BCA), we still found their performance to be indistinguishable from random performance. In other words, even when conditioned on its own belief about whether one item is (relatively) higher-valued than another item, the model is unable to make a rational bet involving the two items.
    \item In investigating RQ3 (Section \ref{bq-2}), we found that the LRMs can make rational decisions when they are fine-tuned on bet questions. If the evaluation modality (say, Coin) is the same as the fine-tuning modality, the performance is typically above 95\%, regardless of the item-set used (i.e., train or test). If the evaluation modality is different from the fine-tuning modality, the LRMs' performance is noticeably lower, but still exceeds 70\% accuracy on average. This suggests that the models are able to achieve (limited) generalization beyond the fine-tuning modality.  
    \item Furthermore, when investigating RQ3 (Section \ref{bq-2}), we also found that re-using the train item-set makes little difference to LRMs' performance (although there is expected overfitting when the train set is completely identical to the test set, including in the choice of modality). The result provides some evidence that the methodology used for constructing bet questions (and more generally, for splitting items into `high-value' and `low-value sets) does not inadvertently lead to item-specific bias in the model. 
    \item Finally, in evaluating RQ2 and RQ3 using the threshold method (introduced in Section \ref{mcqa-ft}) with the three different ground-truths (Strict, Positive Gain, and Non-Negative Gain), we found that the LRMs achieve higher performance, and generalize better, on the easier problem of choosing outcomes with non-negative expected gain rather than outcomes that lead to strictly positive expected gain. Similarly, the models perform even worse (on average) in selecting outcomes that are optimal (Strict) than those that only lead to positive expected gain, but may not necessarily be optimal. 
\end{enumerate}

\section{Discussion}\label{sec:discussion}

Based on the results of RQ1, a natural question arises as to why near-perfect performance was observed on the value questions after fine-tuning the LRMs. Although the LRMs might be `learning' to prefer high-value items over low-value items due to the fine-tuning, we consider this possibility to be unlikely due to the fact that the data set used for fine-tuning is relatively small, and also that the test items are significantly different from the training items (and were independently selected). Rather, the likely reason is that the LRMs are sensitive to \emph{format}, and that fine-tuning the model was akin to teaching it the format and the semantics of the preference being elicited. This allowed it to  generalize to different pairs of (unseen) items, and to learn the correct preference function. While it is certainly possible that the model has learned the same preference function that we used to construct the item sets (i.e., by determining, in a commonsense fashion, whether an item was high-value or low-value), this claim is notoriously difficult to prove due to the black box nature of the LRM.

However, the empirical evidence strongly suggests that, after controlling for format, there is agreement between the LRMs' preference function and ours. Indeed, the near-perfect accuracy on the test set shows that, on average, the LRMs' assignment of high- and low-value items agrees with ours. More importantly, because the LRMs were independently able to replicate our assignment of test items to high- and low-value buckets, errors in the LRMs' performance, including in RQ2, cannot be explained by (hypothetical) arbitrariness in our assignment of high- and low-value items. 

This is further confirmed by the BCA results when evaluating RQ2. Theoretically, if such arbitrariness had existed (or due to some other analogous reason), the BCA \emph{could} have be very different from the ordinary accuracy. One fundamental reason why such a clear alignment was observed between BCA and ordinary accuracy (which, as explained earlier, is due to the agreement between the LRMs', and our, implicit preference functions) may be due to our items being rather extreme in terms of their assigned value. We suspect that BCA will start empirically diverging from standard accuracy when the value difference between items is (arguably) more ambiguous. For example, would an LRM consider a silver ring to be more valuable than a platinum ring (as opposed to, say, a plastic pen, which is a far easier distinction to make)? Conducting similar experiments while controlling for the extremity between the putative values of high- and low-value items is an interesting agenda to consider for future research. 

Considering RQ3, when the fine-tuned LRMs were tested on a modality different from the one used during fine-tuning (`cross-modal setting'), they are more likely to correctly pick outcomes with non-negative expected gain than outcomes with positive expected gain. Although this might be the case due to the former problem seemingly being easier than the latter, there could also be a methodological explanation. Recall that in Section \ref{bq}, we constructed the bet questions in two different ways. One type of question included among its choices an outcome that the bet-maker wins a high-value item, and another outcome that the bet-maker  loses a low-value item. Another type of question had as choices an outcome that the bet-maker  loses a high-value item, and an outcome that the bet-maker wins a low-value item. While we reproduce the mathematical expressions of the expected gain of each outcome for each of the two kinds of bet questions in the Appendix, intuitively, the outcome that maximizes the expected gain of the second type is `do not bet' (which has an expected gain of 0). All other outcomes (in the second type of question) have negative expected gain because, as explained in Section \ref{bq},  the bet-maker  is assumed to place a wager, the value of which lies between the high- and low-value extremes. In contrast, for the first type of question, the bet-maker  should pick the outcome where it can win the high-value item, since this outcome has the highest (strictly positive) expected gain.

Diving deeper into the LRMs' performance across both types of questions, we found that, on average, their performance was higher, and generalization was better, on the second type of question. This further supports the original claim that the models are better able to select outcomes with non-negative expected gain; however, it might also be overfitting on the second type of question by selecting `do not bet' when it recognizes the \emph{type} of the question. In contrast, for the first type of question, the optimal outcome is not \emph{fixed} and overfitting is much more difficult. By way of example, consider the following two questions using the Coin modality:

\begin{enumerate}
    \item If the coin comes up heads, then I win a watch. If it comes up tails, then I lose an egg. What should I do to maximize my expected gains?
    \item If the coin comes up heads, then I lose an egg. If it comes up tails, then I win a watch. What should I do to maximize my expected gains?
\end{enumerate}

Although both instances above are of the first type of question, the optimal outcome in the first is betting on heads, while in the second, the optimal outcome is betting on tails. Hence, unlike in the second type of question, the model does not have the option of `recognizing' the type of question and picking a fixed outcome like `do not bet.' This further underscores the importance of constructing multiple-choice benchmarks that have good internal validity (on the decision-making problem) and minimize the risk of achieving good performance through some form of superficial pattern recognition.

When considering that BERT, RoBERTa, DeBERTa and BigBird each successively build on each other, with performance increases observed for later models across some benchmarks, one might expect to also see similar improvements on decision making. For example, when evaluated on the development set of the Multi-Genre Natural Language Inference-Matched (MNLI-m) benchmark \cite{N18-1101}, these four LRMs achieve accuracy of 88.8\%, 87.6\%, 87.5\%, and 84.6\%, respectively. However, our results show that such improvements are not consistently achieved. This suggests that performance across the decision-making tasks may have a strong dependence on the fundamental (transformer-based) model structure that these LRMs have in common. 

Considering LRMs' cross-modal performance specifically on the test item-sets, it is not evident that there is a \emph{single} modality that we should fine-tune a model on, to consistently achieve the best \emph{aggregate} cross-modal performance. In fact, our results show that there is no one modality that leads to clearly better generalization (defined as using different item-sets and different modalities during testing, compared to fine-tuning). For example, the LRMs were found to achieve the best performance, on average, on the Card modality when they were fine-tuned on the Coin modality. In contrast, they achieved the best aggregate performance on the Dice modality when they were fine-tuned on the Card modality. 

Finally, although three modalities were used in our experiments, there are (potentially) an infinite number of modalities that could be devised for probing these models' decision making. This begs the (open) question of whether there is a `general' fine-tuning procedure that would result in consistently high average performance across any reasonable modality, rather than the one used during fine-tuning.

\section{Conclusion and Future Work}\label{sec:conclusion}

Modern Language Representation Models (LRMs), based on transformer neural networks, have rapidly exceeded the previous state-of-the-art on a range of natural language understanding tasks, including question answering, text summarization, and information extraction \cite{talmor2018commonsenseqa, gliwa2019samsum, kim2003genia}. In this article, we addressed the question of whether such LRMs can be adapted for (approximately) rational decision-making and preference elicitation. In the cognitive science literature, such decision-making is often evaluated using bets. Given the near human-like performance of LRMs on language-based problems, we formulated a set of research questions to specifically test whether: (i) LRMs have distinct preference for high-value items over low-value items, especially when the items were not seen during training, and after stratifying by the format of the questions, (ii) LRMs can make, or be taught to make, (approximately rational) bets in a \emph{generalizable} manner, including when an LRM has been fine-tuned on one `modality' of bet, but is evaluated on another modality.  

We constructed a set of novel benchmarks to empirically test these hypotheses using four established transformer-based LRMs. Our first set of results show that, while LRMs can distinguish between \emph{unseen} high- and low-value items, they  only do so after stratifying by the format of the questions through fine-tuning. This is despite the question being expressed in relatively simple language, and the items being of an everyday nature. The second set of results is similar: LRMs can only make bets (whether posed using the same, or different, modality as the training set) once they have been fine-tuned on similar bet questions. We find, furthermore, that changing the modality of the bet typically leads to a noticeable drop in performance, but is still much higher than random. Thus, while the models do seem to be generalizing, their ability to do so is limited, at best. For a subset of the bet questions, we also find some evidence of overfitting.

There are many promising avenues for future research. Our experiments have only probed the surface of these LRMs' decision-making abilities, since our benchmarks test decision-making when the number of outcomes is limited and equi-probable, and the putative value difference between the pair of items (high- and low-value) is extreme. It remains to be seen whether newer models (such as T5 \cite{raffel2019exploring}) would generalize more effectively in the cross-modal setting, and to more complex decision-making. Considering that there are infinitely many decision-making modalities (in theory), the larger question remains as to the \emph{general} methodology or approach required for the LRMs to achieve human-level decision-making performance on \emph{any} reasonable modality. This question is motivated by the observation that, in the real world, decisions are not framed precisely or explicitly, and the modalities used (if any) are unknown \emph{a priori}. To be applicable in such situations, therefore, LRMs need to be able to make decisions as a \emph{fundamental} capability, as opposed to some form of brute-force fine-tuning on ever larger corpora.

Finally, another direction that could be explored is the use of generative models, such as the Generative Pre-trained Transformer 3 (GPT-3) model \cite{brown2020language}, for decision making. Such models have yielded promising results on zero-shot learning problems, and have even been shown to exhibit human-like creativity. They may be more amenable to decision making in an open-ended setting when no option is given (or can be computed in advance), but an evaluation to that effect has not been conducted yet and is a promising avenue for future research.

\bibliography{mybibfile}

\begin{thebibliography}{10}
\expandafter\ifx\csname url\endcsname\relax
  \def\url#1{\texttt{#1}}\fi
\expandafter\ifx\csname urlprefix\endcsname\relax\def\urlprefix{URL }\fi
\expandafter\ifx\csname href\endcsname\relax
  \def\href#1#2{#2} \def\path#1{#1}\fi

\bibitem{devlin2018bert}
J.~Devlin, M.-W. Chang, K.~Lee, K.~Toutanova, Bert: Pre-training of deep
  bidirectional transformers for language understanding, arXiv preprint
  arXiv:1810.04805.

\bibitem{radford2019language}
A.~Radford, J.~Wu, R.~Child, D.~Luan, D.~Amodei, I.~Sutskever, et~al., Language
  models are unsupervised multitask learners, OpenAI blog 1~(8) (2019) 9.

\bibitem{brown2020language}
T.~Brown, B.~Mann, N.~Ryder, M.~Subbiah, J.~D. Kaplan, P.~Dhariwal,
  A.~Neelakantan, P.~Shyam, G.~Sastry, A.~Askell, et~al., Language models are
  few-shot learners, Advances in neural information processing systems 33
  (2020) 1877--1901.

\bibitem{gupta2022matscibert}
T.~Gupta, M.~Zaki, N.~Krishnan, et~al., Matscibert: A materials domain language
  model for text mining and information extraction, npj Computational Materials
  8~(1) (2022) 1--11.

\bibitem{larionov2019semantic}
D.~Larionov, A.~Shelmanov, E.~Chistova, I.~Smirnov, Semantic role labeling with
  pretrained language models for known and unknown predicates, in: Proceedings
  of the International Conference on Recent Advances in Natural Language
  Processing (RANLP 2019), 2019, pp. 619--628.

\bibitem{aksenov2020abstractive}
D.~Aksenov, J.~Moreno-Schneider, P.~Bourgonje, R.~Schwarzenberg, L.~Hennig,
  G.~Rehm, Abstractive text summarization based on language model conditioning
  and locality modeling, arXiv preprint arXiv:2003.13027.

\bibitem{lample2019cross}
G.~Lample, A.~Conneau, Cross-lingual language model pretraining, arXiv preprint
  arXiv:1901.07291.

\bibitem{izacard2020leveraging}
G.~Izacard, E.~Grave, Leveraging passage retrieval with generative models for
  open domain question answering, arXiv preprint arXiv:2007.01282.

\bibitem{he2020deberta}
P.~He, X.~Liu, J.~Gao, W.~Chen, Deberta: Decoding-enhanced bert with
  disentangled attention, arXiv preprint arXiv:2006.03654.

\bibitem{li2020unimo}
W.~Li, C.~Gao, G.~Niu, X.~Xiao, H.~Liu, J.~Liu, H.~Wu, H.~Wang, Unimo: Towards
  unified-modal understanding and generation via cross-modal contrastive
  learning, arXiv preprint arXiv:2012.15409.

\bibitem{clark2020electra}
K.~Clark, M.-T. Luong, Q.~V. Le, C.~D. Manning, Electra: Pre-training text
  encoders as discriminators rather than generators, arXiv preprint
  arXiv:2003.10555.

\bibitem{khashabi2020unifiedqa}
D.~Khashabi, S.~Min, T.~Khot, A.~Sabharwal, O.~Tafjord, P.~Clark,
  H.~Hajishirzi, Unifiedqa: Crossing format boundaries with a single qa system,
  arXiv preprint arXiv:2005.00700.

\bibitem{leaderboard}
A.~I. for AI, Leaderboard, \url{https://leaderboard.allenai.org/}.

\bibitem{wang2022modern}
Z.~Wang, Modern question answering datasets and benchmarks: A survey, arXiv
  preprint arXiv:2206.15030.

\bibitem{patentbert}
J.-S. Lee, J.~Hsiang, Patentbert: Patent classification with fine-tuning a
  pre-trained bert model, arXiv preprint arXiv:1906.02124.

\bibitem{distilbert}
V.~Sanh, L.~Debut, J.~Chaumond, T.~Wolf, Distilbert, a distilled version of
  bert: smaller, faster, cheaper and lighter, arXiv preprint arXiv:1910.01108.

\bibitem{lee2020biobert}
J.~Lee, W.~Yoon, S.~Kim, D.~Kim, S.~Kim, C.~H. So, J.~Kang, Biobert: a
  pre-trained biomedical language representation model for biomedical text
  mining, Bioinformatics 36~(4) (2020) 1234--1240.

\bibitem{docbert}
A.~Adhikari, A.~Ram, R.~Tang, J.~Lin, Docbert: Bert for document
  classification, arXiv preprint arXiv:1904.08398.

\bibitem{kbert}
W.~Liu, P.~Zhou, Z.~Zhao, Z.~Wang, Q.~Ju, H.~Deng, P.~Wang, K-bert: Enabling
  language representation with knowledge graph, in: Proceedings of the AAAI
  Conference on Artificial Intelligence, Vol.~34, 2020, pp. 2901--2908.

\bibitem{scibert}
I.~Beltagy, K.~Lo, A.~Cohan, Scibert: A pretrained language model for
  scientific text, arXiv preprint arXiv:1903.10676.

\bibitem{sun2019videobert}
C.~Sun, A.~Myers, C.~Vondrick, K.~Murphy, C.~Schmid, Videobert: A joint model
  for video and language representation learning, in: Proceedings of the
  IEEE/CVF International Conference on Computer Vision, 2019, pp. 7464--7473.

\bibitem{lu2019vilbert}
J.~Lu, D.~Batra, D.~Parikh, S.~Lee, Vilbert: Pretraining task-agnostic
  visiolinguistic representations for vision-and-language tasks, arXiv preprint
  arXiv:1908.02265.

\bibitem{beckage2016language}
N.~M. Beckage, E.~Colunga, Language networks as models of cognition:
  Understanding cognition through language, Towards a theoretical framework for
  analyzing complex linguistic networks (2016) 3--28.

\bibitem{harris2006language}
C.~L. Harris, Language and cognition, Encyclopedia of cognitive science (2006)
  1--6.

\bibitem{wallace2019nlp}
E.~Wallace, Y.~Wang, S.~Li, S.~Singh, M.~Gardner, Do nlp models know numbers?
  probing numeracy in embeddings, arXiv preprint arXiv:1909.07940.

\bibitem{balasubramanian2020s}
S.~Balasubramanian, N.~Jain, G.~Jindal, A.~Awasthi, S.~Sarawagi, What's in a
  name? are bert named entity representations just as good for any other name?,
  arXiv preprint arXiv:2007.06897.

\bibitem{industry1}
T.-T. Dang, T.-B. Ho, Mixture of language models utilization in score-based
  sentiment classification on clinical narratives, in: International Conference
  on Industrial, Engineering and Other Applications of Applied Intelligent
  Systems, Springer, 2016, pp. 255--268.

\bibitem{industry2}
P.~P. Bonissone, What is new in industry?[industrial and governmental
  activities], IEEE Computational Intelligence Magazine 17~(3) (2022) 5--6.

\bibitem{industry3}
H.~Zhu, P.~Tiwari, A.~Ghoneim, M.~S. Hossain, A collaborative ai-enabled
  pretrained language model for aiot domain question answering, IEEE
  Transactions on Industrial Informatics 18~(5) (2021) 3387--3396.

\bibitem{industry4}
M.~Shoeybi, M.~Patwary, R.~Puri, P.~LeGresley, J.~Casper, B.~Catanzaro,
  Megatron-lm: Training multi-billion parameter language models using model
  parallelism, arXiv preprint arXiv:1909.08053.

\bibitem{wei2022emergent}
J.~Wei, Y.~Tay, R.~Bommasani, C.~Raffel, B.~Zoph, S.~Borgeaud, D.~Yogatama,
  M.~Bosma, D.~Zhou, D.~Metzler, et~al., Emergent abilities of large language
  models, arXiv preprint arXiv:2206.07682.

\bibitem{kaplan2020scaling}
J.~Kaplan, S.~McCandlish, T.~Henighan, T.~B. Brown, B.~Chess, R.~Child,
  S.~Gray, A.~Radford, J.~Wu, D.~Amodei, Scaling laws for neural language
  models, arXiv preprint arXiv:2001.08361.

\bibitem{wang2021measure}
X.~Wang, H.~Wang, D.~Yang, Measure and improve robustness in nlp models: A
  survey, arXiv preprint arXiv:2112.08313.

\bibitem{fursov2022differentiable}
I.~Fursov, A.~Zaytsev, P.~Burnyshev, E.~Dmitrieva, N.~Klyuchnikov,
  A.~Kravchenko, E.~Artemova, E.~Komleva, E.~Burnaev, A differentiable language
  model adversarial attack on text classifiers, IEEE Access 10 (2022)
  17966--17976.

\bibitem{heinzerling2020language}
B.~Heinzerling, K.~Inui, Language models as knowledge bases: On entity
  representations, storage capacity, and paraphrased queries, arXiv preprint
  arXiv:2008.09036.

\bibitem{bscience1}
C.~Camerer, Bounded rationality in individual decision making, Experimental
  economics 1~(2) (1998) 163--183.

\bibitem{bscience2}
C.~G. Hempel, Valuation and objectivity in science, in: A Portrait of
  Twenty-five Years, Springer, 1983, pp. 277--304.

\bibitem{bscience3}
J.-P. Beno{\^\i}t, J.~Dubra, D.~A. Moore, Does the better-than-average effect
  show that people are overconfident?: Two experiments, Journal of the European
  Economic Association 13~(2) (2015) 293--329.

\bibitem{sonsino2002rationality}
D.~Sonsino, I.~Erev, S.~Gilat, On rationality, learning and zero-sum
  betting—an experimental study of the no-betting conjecture, Retrieved
  January 12 (2002) 2017.

\bibitem{kahneman1979d}
T.~Kahneman, D. kahneman, a. tversky, Prospect theory: An analysis of decisions
  under risk (1979) 263--291.

\bibitem{kahneman2013prospect}
D.~Kahneman, A.~Tversky, Prospect theory: An analysis of decision under risk,
  in: Handbook of the fundamentals of financial decision making: Part I, World
  Scientific, 2013, pp. 99--127.

\bibitem{du2022shortcut}
M.~Du, F.~He, N.~Zou, D.~Tao, X.~Hu, Shortcut learning of large language models
  in natural language understanding: A survey, arXiv preprint arXiv:2208.11857.

\bibitem{ke3}
K.~Shen, M.~Kejriwal, On the generalization abilities of fine-tuned commonsense
  language representation models, in: International Conference on Innovative
  Techniques and Applications of Artificial Intelligence, Springer, 2021, pp.
  3--16.

\bibitem{tommasi2017deeper}
T.~Tommasi, N.~Patricia, B.~Caputo, T.~Tuytelaars, A deeper look at dataset
  bias, in: Domain adaptation in computer vision applications, Springer, 2017,
  pp. 37--55.

\bibitem{wang2018glue}
A.~Wang, A.~Singh, J.~Michael, F.~Hill, O.~Levy, S.~R. Bowman, Glue: A
  multi-task benchmark and analysis platform for natural language
  understanding, arXiv preprint arXiv:1804.07461.

\bibitem{rajpurkar2016squad}
P.~Rajpurkar, J.~Zhang, K.~Lopyrev, P.~Liang, Squad: 100,000+ questions for
  machine comprehension of text, arXiv preprint arXiv:1606.05250.

\bibitem{rajpurkar2018know}
P.~Rajpurkar, R.~Jia, P.~Liang, Know what you don't know: Unanswerable
  questions for squad, arXiv preprint arXiv:1806.03822.

\bibitem{liu2019roberta}
Y.~Liu, M.~Ott, N.~Goyal, J.~Du, M.~Joshi, D.~Chen, O.~Levy, M.~Lewis,
  L.~Zettlemoyer, V.~Stoyanov, Roberta: A robustly optimized bert pretraining
  approach, arXiv preprint arXiv:1907.11692.

\bibitem{williams2017broad}
A.~Williams, N.~Nangia, S.~R. Bowman, A broad-coverage challenge corpus for
  sentence understanding through inference, arXiv preprint arXiv:1704.05426.

\bibitem{lai2017race}
G.~Lai, Q.~Xie, H.~Liu, Y.~Yang, E.~Hovy, Race: Large-scale reading
  comprehension dataset from examinations, arXiv preprint arXiv:1704.04683.

\bibitem{zaheer2020big}
M.~Zaheer, G.~Guruganesh, K.~A. Dubey, J.~Ainslie, C.~Alberti, S.~Ontanon,
  P.~Pham, A.~Ravula, Q.~Wang, L.~Yang, et~al., Big bird: Transformers for
  longer sequences, Advances in Neural Information Processing Systems 33 (2020)
  17283--17297.

\bibitem{thoppilan2022lamda}
R.~Thoppilan, D.~De~Freitas, J.~Hall, N.~Shazeer, A.~Kulshreshtha, H.-T. Cheng,
  A.~Jin, T.~Bos, L.~Baker, Y.~Du, et~al., Lamda: Language models for dialog
  applications, arXiv preprint arXiv:2201.08239.

\bibitem{chowdhery2022palm}
A.~Chowdhery, S.~Narang, J.~Devlin, M.~Bosma, G.~Mishra, A.~Roberts, P.~Barham,
  H.~W. Chung, C.~Sutton, S.~Gehrmann, et~al., Palm: Scaling language modeling
  with pathways, arXiv preprint arXiv:2204.02311.

\bibitem{BERTGoogle}
P.~Nayak,
  \href{https://blog.google/products/search/search-language-understanding-bert/}{Understanding
  searches better than ever before} (2019).
\newline\urlprefix\url{https://blog.google/products/search/search-language-understanding-bert/}

\bibitem{rogers2020primer}
A.~Rogers, O.~Kovaleva, A.~Rumshisky, A primer in bertology: What we know about
  how bert works, Transactions of the Association for Computational Linguistics
  8 (2020) 842--866.

\bibitem{wu2019mask}
X.~Wu, T.~Zhang, L.~Zang, J.~Han, S.~Hu, " mask and infill": Applying masked
  language model to sentiment transfer, arXiv preprint arXiv:1908.08039.

\bibitem{liu2019linguistic}
N.~F. Liu, M.~Gardner, Y.~Belinkov, M.~E. Peters, N.~A. Smith, Linguistic
  knowledge and transferability of contextual representations, arXiv preprint
  arXiv:1903.08855.

\bibitem{warstadt2020can}
A.~Warstadt, S.~R. Bowman, Can neural networks acquire a structural bias from
  raw linguistic data?, arXiv preprint arXiv:2007.06761.

\bibitem{kobayashi2020attention}
G.~Kobayashi, T.~Kuribayashi, S.~Yokoi, K.~Inui, Attention module is not only a
  weight: Analyzing transformers with vector norms, arXiv preprint
  arXiv:2004.10102.

\bibitem{ettinger2020bert}
A.~Ettinger, What bert is not: Lessons from a new suite of psycholinguistic
  diagnostics for language models, Transactions of the Association for
  Computational Linguistics 8 (2020) 34--48.

\bibitem{ribeiro2020beyond}
M.~T. Ribeiro, T.~Wu, C.~Guestrin, S.~Singh, Beyond accuracy: Behavioral
  testing of nlp models with checklist, arXiv preprint arXiv:2005.04118.

\bibitem{jawahar2019does}
G.~Jawahar, B.~Sagot, D.~Seddah, What does bert learn about the structure of
  language?, in: ACL 2019-57th Annual Meeting of the Association for
  Computational Linguistics, 2019.

\bibitem{wu2020perturbed}
Z.~Wu, Y.~Chen, B.~Kao, Q.~Liu, Perturbed masking: Parameter-free probing for
  analyzing and interpreting bert, arXiv preprint arXiv:2004.14786.

\bibitem{ke1}
M.~Kejriwal, K.~Shen, Do fine-tuned commonsense language models really
  generalize?, arXiv preprint arXiv:2011.09159.

\bibitem{ke2}
K.~Shen, M.~Kejriwal, Understanding prior bias and choice paralysis in
  transformer-based language representation models through four experimental
  probes, arXiv preprint arXiv:2210.01258.

\bibitem{misra2022semantic}
K.~Misra, On semantic cognition, inductive generalization, and language models,
  in: Proceedings of the AAAI Conference on Artificial Intelligence, Vol.~36,
  2022, pp. 12894--12895.

\bibitem{li2022pre}
S.~Li, X.~Puig, Y.~Du, C.~Wang, E.~Akyurek, A.~Torralba, J.~Andreas,
  I.~Mordatch, Pre-trained language models for interactive decision-making,
  arXiv preprint arXiv:2202.01771.

\bibitem{misra2021language}
K.~Misra, A.~Ettinger, J.~T. Rayz, Do language models learn typicality
  judgments from text?, arXiv preprint arXiv:2105.02987.

\bibitem{sawayama2022watching}
M.~Sawayama, Y.~Lemesle, P.-Y. Oudeyer, Watching artificial intelligence
  through the lens of cognitive science methodologies.

\bibitem{zhao2021calibrate}
Z.~Zhao, E.~Wallace, S.~Feng, D.~Klein, S.~Singh, Calibrate before use:
  Improving few-shot performance of language models, in: International
  Conference on Machine Learning, PMLR, 2021, pp. 12697--12706.

\bibitem{vaswani2017attention}
A.~Vaswani, N.~Shazeer, N.~Parmar, J.~Uszkoreit, L.~Jones, A.~N. Gomez,
  {\L}.~Kaiser, I.~Polosukhin, Attention is all you need, Advances in neural
  information processing systems 30.

\bibitem{joshi2017triviaqa}
M.~Joshi, E.~Choi, D.~S. Weld, L.~Zettlemoyer, Triviaqa: A large scale
  distantly supervised challenge dataset for reading comprehension, arXiv
  preprint arXiv:1705.03551.

\bibitem{zellers2018swag}
R.~Zellers, Y.~Bisk, R.~Schwartz, Y.~Choi, Swag: A large-scale adversarial
  dataset for grounded commonsense inference, arXiv preprint arXiv:1808.05326.

\bibitem{yang2018hotpotqa}
Z.~Yang, P.~Qi, S.~Zhang, Y.~Bengio, W.~W. Cohen, R.~Salakhutdinov, C.~D.
  Manning, Hotpotqa: A dataset for diverse, explainable multi-hop question
  answering, arXiv preprint arXiv:1809.09600.

\bibitem{tiedemann2016finding}
J.~Tiedemann, Finding alternative translations in a large corpus of movie
  subtitle, in: Proceedings of the Tenth International Conference on Language
  Resources and Evaluation (LREC'16), 2016, pp. 3518--3522.

\bibitem{Zafarani+Liu:2009}
R.~Zafarani, H.~Liu, \href{http://socialcomputing.asu.edu}{Social computing
  data repository at {ASU}} (2009).
\newline\urlprefix\url{http://socialcomputing.asu.edu}

\bibitem{AWSBERT}
C.~Anastasiu, H.~Behnke, S.~L{\"u}ck, V.~Malesevic, A.~Najmi, J.~Poveda-Panter,
  Deeptitle--leveraging bert to generate search engine optimized headlines,
  arXiv preprint arXiv:2107.10935.

\bibitem{beltagy2019scibert}
I.~Beltagy, K.~Lo, A.~Cohan, Scibert: A pretrained language model for
  scientific text, arXiv preprint arXiv:1903.10676.

\bibitem{lan2019albert}
Z.~Lan, M.~Chen, S.~Goodman, K.~Gimpel, P.~Sharma, R.~Soricut, Albert: A lite
  bert for self-supervised learning of language representations, arXiv preprint
  arXiv:1909.11942.

\bibitem{Fan2007ASO}
R.-E. Fan, C.-J. Lin, A study on threshold selection for multi-label
  classification, 2007.

\bibitem{liu2019knowledge}
A.~Liu, J.~Du, V.~Stoyanov, Knowledge-augmented language model and its
  application to unsupervised named-entity recognition, arXiv preprint
  arXiv:1904.04458.

\bibitem{hfbert}
Hugging face repository: bert-base-uncased,
  \url{https://huggingface.co/bert-base-uncased}.

\bibitem{hfroberta}
Hugging face repository: roberta-base,
  \url{https://huggingface.co/roberta-base}.

\bibitem{hfdeberta}
Hugging face repository: deberta-base,
  \url{https://huggingface.co/microsoft/deberta-base}.

\bibitem{hfbigbird}
Hugging face repository: bigbird-roberta-base,
  \url{https://huggingface.co/google/bigbird-roberta-base}.

\bibitem{tversky1974judgment}
A.~Tversky, D.~Kahneman, Judgment under uncertainty: Heuristics and biases:
  Biases in judgments reveal some heuristics of thinking under uncertainty.,
  science 185~(4157) (1974) 1124--1131.

\bibitem{jin2020bert}
D.~Jin, Z.~Jin, J.~T. Zhou, P.~Szolovits, Is bert really robust? a strong
  baseline for natural language attack on text classification and entailment,
  in: Proceedings of the AAAI conference on artificial intelligence, Vol.~34,
  2020, pp. 8018--8025.

\bibitem{hendrycks2021measuring}
D.~Hendrycks, C.~Burns, S.~Kadavath, A.~Arora, S.~Basart, E.~Tang, D.~Song,
  J.~Steinhardt, Measuring mathematical problem solving with the math dataset,
  arXiv preprint arXiv:2103.03874.

\bibitem{lin2020birds}
B.~Y. Lin, S.~Lee, R.~Khanna, X.~Ren, Birds have four legs?! numersense:
  Probing numerical commonsense knowledge of pre-trained language models, arXiv
  preprint arXiv:2005.00683.

\bibitem{N18-1101}
A.~Williams, N.~Nangia, S.~Bowman,
  \href{http://aclweb.org/anthology/N18-1101}{A broad-coverage challenge corpus
  for sentence understanding through inference}, in: Proceedings of the 2018
  Conference of the North American Chapter of the Association for Computational
  Linguistics: Human Language Technologies, Volume 1 (Long Papers), Association
  for Computational Linguistics, 2018, pp. 1112--1122.
\newline\urlprefix\url{http://aclweb.org/anthology/N18-1101}

\bibitem{talmor2018commonsenseqa}
A.~Talmor, J.~Herzig, N.~Lourie, J.~Berant, Commonsenseqa: A question answering
  challenge targeting commonsense knowledge, arXiv preprint arXiv:1811.00937.

\bibitem{gliwa2019samsum}
B.~Gliwa, I.~Mochol, M.~Biesek, A.~Wawer, Samsum corpus: A human-annotated
  dialogue dataset for abstractive summarization, arXiv preprint
  arXiv:1911.12237.

\bibitem{kim2003genia}
J.-D. Kim, T.~Ohta, Y.~Tateisi, J.~Tsujii, Genia corpus—a semantically
  annotated corpus for bio-textmining, Bioinformatics 19~(suppl\_1) (2003)
  i180--i182.

\bibitem{raffel2019exploring}
C.~Raffel, N.~Shazeer, A.~Roberts, K.~Lee, S.~Narang, M.~Matena, Y.~Zhou,
  W.~Li, P.~J. Liu, Exploring the limits of transfer learning with a unified
  text-to-text transformer, arXiv preprint arXiv:1910.10683.

\end{thebibliography}
\section*{Appendix}

\subsection*{Analysis of expected gain of the bet outcomes}
\label{analysis_bet}
\begin{table}[H]
\resizebox{\columnwidth}{!}{%
\begin{tabular}{ cc }
    \hline
    Choice & Expected Gain \\
    \hline
    Head & $0.5*(H+X)+0.5*0-X=0.5*(H-X)$ \\
    Tail & $0.5*0+0.5*(-L+X)-X=0.5*(-L-X)$ \\
    Head and Tail & $0.5*0.5*(H-X) + 0.5*0.5(-L-X)= 0.5*(0.5H-0.5L-X)$ \\
    Do Not Bet & $X-X=0$ \\
    \hline
\end{tabular}%
}
\caption{The expected gain of bet outcomes corresponding to winning a high value item, and losing a low value item. We use the Coin modality as an example here. The prompt is: \emph{If the coin comes up heads, then I win a `high value item'. If it comes up tails, then I lose a `low value item'. What should I do to maximize my expected gains?} Here, X represents the bet wager amount, while H and L represent the expected monetary value of the high value item and low value item, respectively. The choice `Head and Tail' represents the the situation where the model selects both the `Head' and `Tail' as its (multi-label) prediction, in which case X is assumed to be equally split and wagered on both heads (X/2) and tails (X/2).}
\end{table}

\begin{table}[H]
\resizebox{\columnwidth}{!}{%
\begin{tabular}{ cc }
    \hline
    Choice & Expected Gain \\
    \hline
    Head & $0.5*(L+X) + 0.5*0 -X=0.5*(L-X)$ \\
    Tail & $0.5*0+0.5*(-H+X)-X=0.5*(-H-X)$ \\
    Head and Tail & $0.5*0.5*(L-X) + 0.5*0.5(-H-X)= 0.5*(0.5*L-0.5H-X)$ \\
    Do Not Bet & $X-X=0$ \\
    \hline
\end{tabular}%
}
\caption{The expected gain of bet outcomes corresponding to losing a high value item, and winning a low value item. We use the Coin modality as an example here. The prompt is: \emph{If the coin comes up heads, then I lose a `high value item'. If it comes up tails, then I win a `low value item'. What should I do to maximize my expected gains?} Here, X represents the bet wager amount, while H and L represent the expected monetary value of the high value item and low value item, respectively. The choice `Head and Tail' represents the the situation where the model selects both the `Head' and `Tail' as its (multi-label) prediction, in which case X is assumed to be equally split and wagered  on both heads (X/2) and tails (X/2).}
\end{table}

\subsection*{P values for RQ1}
\label{urlqr1p}

\begin{table}[H]
\centering
\begin{tabular}{ ccccc }
 \hline
 Template $\rightarrow$ &BE & BV & CE& CV\\
 $\downarrow$ Model& \multicolumn{4}{c}{Default/ Fine-tuned} \\
 \hline
 BERT    & $<$.001/$<$.001 &.392/$<$.001 &.013/$<$.001 &.033/$<$.001 \\
 RoBERTa & .857/$<$.001 &.075/$<$.001 &.146/$<$.001 &.146/$<$.001 \\
 DeBERTa & 1.00/$<$.001 & 1.00/$<$.001 &.252/$<$.001 &.075/$<$.001 \\
 BigBird & 1.00/$<$.001 & 1.00/$<$.001 &.033/$<$.001 &.075/$<$.001 \\
 \hline
\end{tabular}
\caption{P values corresponding to the results in Table \ref{table:rq1-result}, when comparing to random performance of 33\%.}
\label{table:rq1-result-pvalue}
\end{table}

\begin{table}[H]
\centering
\begin{tabular}{ ccccc }
 \hline
 Template $\rightarrow$ &BE & BV & CE& CV\\
 $\downarrow$ Model& \multicolumn{4}{c}{Default/ Fine-tuned} \\
 \hline
 BERT    & .792/$<$.001 & 1.00/$<$.001 & .319/$<$.001 & .530/$<$.001 \\
 RoBERTa & .983/$<$.001 & .983/$<$.001 & 1.00/$<$.001 & 1.00/$<$.001 \\
 DeBERTa & 1.00/$<$.001 & 1.00/$<$.001 & 1.00/$<$.001 & 1.00/$<$.001 \\
 BigBird & 1.00/$<$.001 & 1.00/$<$.001 & .179/$<$.001 & .093/$<$.001 \\
 \hline
\end{tabular}
\caption{P values corresponding to the `Normal' threshold-based results in Figure \ref{figure:rq1}, when comparing to random performance of 12.5\%.}
\end{table}

\begin{table}[H]
\centering
\begin{tabular}{ ccccc }
 \hline
 Template $\rightarrow$ &BE & BV & CE& CV\\
 $\downarrow$ Model& \multicolumn{4}{c}{Default/ Fine-tuned} \\
 \hline
 BERT    & 1.00/$<$.001 & 1.00/$<$.001 & .885/$<$.001 & .974/$<$.001 \\
 RoBERTa & .974/$<$.001 & .885/$<$.001 & 1.00/$<$.001 & 1.00/$<$.001 \\
 DeBERTa & 1.00/$<$.001 & 1.00/$<$.001 & 1.00/$<$.001 & 1.00/$<$.001 \\
 BigBird & 1.00/$<$.001 & 1.00/$<$.001 & .729/$<$.001 & .545/$<$.001 \\
 \hline
\end{tabular}
\caption{P values corresponding to the `Weak Normal' threshold-based results in Figure \ref{figure:rq1}, when comparing to random performance of 25\%.}
\end{table}

\begin{table}[H]
\centering
\begin{tabular}{ ccccc }
 \hline
 Template $\rightarrow$ &BE & BV & CE& CV\\
 $\downarrow$ Model& \multicolumn{4}{c}{Default/ Fine-tuned} \\
 \hline
 BERT    & 1.00/.281 & 1.00/.061 & .848/$<$.001 & .922/$<$.001 \\
 RoBERTa & 1.00/.061 & 1.00/.149 & .848/$<$.001 & .439/$<$.001 \\
 DeBERTa & .966/.061 & .987/.439 & 1.00/$<$.001 & 1.00/$<$.001 \\
 BigBird & $<$.001/$<$.001 & .149/$<$.001 & 1.00/$<$.001 & 1.00/$<$.001 \\
 \hline
\end{tabular}
\caption{P values corresponding to the `Weak' threshold-based results in Figure \ref{figure:rq1}, when comparing to random performance of 62.5\%.}
\end{table}

\subsection*{P values for RQ2}

\begin{table}[H]
\centering

\resizebox{\columnwidth}{!}{%
\begin{tabular}{ ccccccccc }
 \hline
 Modality $\rightarrow$ & \multicolumn{2}{c}{Coin} && \multicolumn{2}{c}{Dice} && \multicolumn{2}{c}{Card} \\
 \hline
 Metric $\rightarrow$ & ACC & BCA & & ACC & BCA && ACC & BCA \\ 
 $\downarrow$ Model& \multicolumn{8}{c}{Default/Fine-tuned} \\
 \hline
 BERT & .026/$<$.001 &.040/$<$.001 && .972/$<$.001 & .972/$<$.001 && .040/$<$.001 & .922/$<$.001\\
 RoBERTa & .972/.972 & .972/.972 && .003/.972 &.026/.972 && .829/.612 & .881/.612\\
 DeBERTa & $<$.001/.016 & $<$.001/.016 && $<$.001/$<$.001 & .009/$<$.001 && .002/.003 & .003/.003\\
 BigBird & .364/.972 & .124/.972 && .002/.828 & .006/.828 && .225/.001 & .364/.001\\
 \hline
\end{tabular}%
}
\caption{P values corresponding to the results in Table \ref{table:rq2-result-raw}, when comparing to random performance of 33\%.}
\label{rq2-pvalue}
\end{table}

\subsection*{Threshold method results for RQ2}

Table \ref{table:rq2-result-threshold} reports the LRMs' results on the three bet modalities using the threshold method detailed in Section \ref{mcqa-ft}, using the three different ground truths discussed in Section \ref{bq-1}.

\begin{table}[H]
\caption{LRMs' performance (expressed as percentage) for instantiated bet questions, defined in Table \ref{table:bet-question}, using the threshold method for each of three different ground-truths (Strict, Positive Gain, Non-Negative Gain) discussed in Section \ref{bq}. Where fine-tuning is involved, the LRMs are fine-tuned on (Choice Valuable) value questions introduced in Table \ref{table:value-question}. $\S$ indicates that the result is statistically better than random performance with 95\% confidence.}
\begin{tabular}{ ccccc }
 \hline
 Modality $\rightarrow$ & & Coin & Dice & Card \\
 \hline
 Model &  Ground-truth & \multicolumn{3}{c}{Default/Fine-tuned} \\
 \hline
 \multirow{3}{*}{BERT} & Strict & 0/0 & 0/0 & 0/12 \\
 & Positive Gain & 0/0 & 0/0 & 0/0 \\
 & Non-Negative Gain & 0/0 & 0/0 & 3/22 \\
 \hline
 \multirow{3}{*}{RoBERTa} & Strict & 0/0 & 0/0 & 0/5 \\
 & Positive Gain & 0/2 & 0/0 & 0/6\\
 & Non-Negative Gain & 0/1 & 0/0 & 0/3\\
 \hline
 \multirow{3}{*}{DeBERTa} & Strict & 0/0 & 0/0 & 0/0 \\
 & Positive Gain & 0/0 & 0/0 & 0/0\\
 & Non-Negative Gain & 0/0 & 0/0 & 0/0 \\
 \hline
 \multirow{3}{*}{BigBird} & Strict &0/0  & 0/0 & 0/25$^\S$ \\
 & Positive Gain & 0/0 & 0/0 & 0/0 \\
 & Non-Negative Gain & 0/0 & 0/0 & 0/52$^\S$ \\
 \hline
\end{tabular}
\label{table:rq2-result-threshold}
\end{table}

The threshold results confirm that the LRMs cannot correctly choose the option that maximize the expected gain (Strict), or even achieves positive expected gain (Positive Gain). The best performance, among these two ground-truths, is only 25\%, achieved by the fine-tuned BigBird using the Strict ground-truth. However, we do observe some improvements when evaluating the LRMs using the Non-Negative Gain ground-truth e.g., in this setting, fine-tuned BigBird is able to achieve 52\%. This improvement in performance suggests that selecting options that have non-negative expected gain may be a easier problem for the LRMs than selecting options with strictly positive, and optimal, gains.

Moreover, we observe some general improvements when we compare the results of the default models to those of the fine-tuned models. This suggests that, using the threshold method, the fine-tuned LRMs can perform slightly better than their default counterparts, but are still not able to significantly outperform random selection. Interestingly, we observe that LRMs' consistently achieve better performance on the Card modality than on the other two modalities, which suggests that the Card modality might be a `easier' modality (at least for the LRMs) than the other two. In general, the conclusion from the threshold-based results is in alignment with the previous conclusion that neither  the default nor (value questions) fine-tuned  LRMs perform better than random in making rational bets.

The P values corresponding to the results in Table \ref{table:rq2-result-threshold} are tabulated in Table \ref{pvalues:thresholdRQ2}.

\begin{table}[H]
\centering
\caption{P values corresponding to the results in Table \ref{table:rq2-result-threshold}}
\begin{tabular}{ ccccc }
 \hline
 Modality $\rightarrow$ & & Coin & Dice & Card \\
 \hline
 $\downarrow$ Model& $\downarrow$ Ground-truth & \multicolumn{3}{c}{Default/Fine-tuned} \\
 \hline
 \multirow{3}{*}{BERT} & Strict & 1.00/1.00 & 1.00/1.00 & 1.00/.561 \\
 & Positive Gain & 1.00/1.00 & 1.00/1.00 & 1.00/1.00 \\
 & Non-Negative Gain & 1.00/1.00 & 1.00/1.00 & 1.00/.764 \\
 \hline
 \multirow{3}{*}{RoBERTa} & Strict & 1.00/1.00 & 1.00/1.00 & 1.00/1.00 \\
 & Positive Gain & 1.00/1.00 & 1.00/1.00 & 1.00/1.00 \\
 & Non-Negative Gain & 1.00/1.00 & 1.00/1.00 & 1.00/1.00 \\
 \hline
 \multirow{3}{*}{DeBERTa} & Strict & 1.00/1.00 & 1.00/1.00 & 1.00/1.00 \\
 & Positive Gain & 1.00/1.00 & 1.00/1.00 & 1.00/1.00 \\
 & Non-Negative Gain & 1.00/1.00 & 1.00/1.00 & 1.00/1.00 \\
 \hline
 \multirow{3}{*}{BigBird} & Strict & 1.00/1.00 & 1.00/1.00 & 1.00/.002 \\
 & Positive Gain & 1.00/1.00 & 1.00/1.00 & 1.00/1.00 \\
 & Non-Negative Gain & 1.00/1.00 & 1.00/1.00 & 1.00/$<$.001 \\
 \hline
\end{tabular}\label{pvalues:thresholdRQ2}
\end{table}

\subsection*{P values for RQ3}

\begin{table}[H]
\centering
\resizebox{\columnwidth}{!}{%
\begin{tabular}{ cccccccccccc }
 \hline
 Fine-tune on $\rightarrow$ & Card &&&& Coin &&&& Dice &&\\
 Test on $\rightarrow$ & Card & Coin & Dice && Card & Coin & Dice && Card & Coin & Dice \\ 
 $\downarrow$ Model& \multicolumn{11}{c}{Train Set/Test Set} \\
 \hline
 BERT & $<$.001/$<$.001 & $<$.001/$<$.001 & $<$.001/$<$.001 && $<$.001/$<$.001 & $<$.001/$<$.001 & $<$.001/$<$.001 && $<$.001/$<$.001 & $<$.001/$<$.001 & $<$.001/$<$.001 \\
 RoBERTa & $<$.001/$<$.001 & $<$.001/$<$.001 & $<$.001/$<$.001 && $<$.001/$<$.001 & $<$.001/$<$.001 & $<$.001/.004 && $<$.001/$<$.001 & $<$.001/$<$.001 & $<$.001/$<$.001 \\
 DeBERTa & $<$.001/$<$.001 & $<$.001/$<$.001 & $<$.001/$<$.001 && $<$.001/$<$.001 & $<$.001/$<$.001 & $<$.001/$<$.001 && $<$.001/$<$.001 & $<$.001/$<$.001 & $<$.001/$<$.001 \\
 BigBird & $<$.001/$<$.001 & $<$.001/$<$.001 & $<$.001/$<$.001 && $<$.001/$<$.001 & $<$.001/$<$.001 & $<$.001/$<$.001 && $<$.001/$<$.001 & $<$.001/$<$.001 & $<$.001/$<$.001 \\
 \hline
\end{tabular}%
}
\caption{P values corresponding to the results in Figure \ref{table:rq2-result-fig}, when comparing to random performance of 33\%.}
\end{table}

\begin{table}[H]
\centering
\resizebox{\columnwidth}{!}{%
\begin{tabular}{ ccccccccccccc }
 \hline
 \multicolumn{2}{c}{Fine-tune on} $\rightarrow$ & Card &&&& Coin &&&& Dice &&\\
 $\downarrow$ Model & Test on $\rightarrow$ & Card & Coin & Dice && Card & Coin & Dice && Card & Coin & Dice \\
 \hline
 \multicolumn{2}{c}{BERT} & $<$.001 & $<$.001 & .011 && $<$.001 & $<$.001 & $<$.001 && $<$.001 & $<$.001 & $<$.001 \\
 \multicolumn{2}{c}{RoBERTa} & $<$.001 & $<$.001 & $<$.001 && $<$.001 & $<$.001 & $<$.001 && $<$.001 & $<$.001 & $<$.001 \\
 \multicolumn{2}{c}{DeBERTa} & $<$.001 & $<$.001 & $<$.001 && $<$.001 & $<$.001 & .011 && $<$.001 & $<$.001 & $<$.001 \\
 \multicolumn{2}{c}{BigBird} & $<$.001 & $<$.001 & $<$.001 && $<$.001 & $<$.001 & $<$.001 && $<$.001 & $<$.001 & $<$.001 \\
 \hline
\end{tabular}%
}
\caption{P values corresponding to the `Strict' results in Figure \ref{table:rq3-threshold}, when comparing to random performance of 12.5\%.}
\end{table}

\begin{table}[H]
\centering
\resizebox{\columnwidth}{!}{%
\begin{tabular}{ ccccccccccccc }
 \hline
 \multicolumn{2}{c}{Fine-tune on} $\rightarrow$ & Card &&&& Coin &&&& Dice &&\\
 $\downarrow$ Model & Test on $\rightarrow$ & Card & Coin & Dice && Card & Coin & Dice && Card & Coin & Dice \\
 \hline
 \multicolumn{2}{c}{BERT} & $<$.001 & 1.00 & 1.00 && $<$.001 & $<$.001 & $<$.001 && $<$.001 & $<$.001 & $<$.001 \\
 \multicolumn{2}{c}{RoBERTa} & $<$.001 & $<$.001 & $<$.001 && $<$.001 & $<$.001 & $<$.001 && $<$.001 & $<$.001 & $<$.001 \\
 \multicolumn{2}{c}{DeBERTa} & $<$.001 & $<$.001 & $<$.001 && $<$.001 & $<$.001 & $<$.001 && $<$.001 & $<$.001 & $<$.001 \\
 \multicolumn{2}{c}{BigBird} & $<$.001 & 1.00 & .437 && 1.00 & $<$.001 & $<$.001 && .222 & .222 & $<$.001 \\
 \hline
\end{tabular}%
}
\caption{P values corresponding to the `Positive Gain' results in Figure \ref{table:rq3-threshold}, when comparing to random performance of 25\%.}
\end{table}

\begin{table}[H]
\centering
\resizebox{\columnwidth}{!}{%
\begin{tabular}{ ccccccccccccc }
 \hline
 \multicolumn{2}{c}{Fine-tune on} $\rightarrow$ & Card &&&& Coin &&&& Dice &&\\
 $\downarrow$ Model & Test on $\rightarrow$ & Card & Coin & Dice && Card & Coin & Dice && Card & Coin & Dice \\
 \hline
 \multicolumn{2}{c}{BERT} & $<$.001 & $<$.001 & .764 && $<$.001 & $<$.001 & $<$.001 && $<$.001 & $<$.001 & $<$.001 \\
 \multicolumn{2}{c}{RoBERTa} & $<$.001 & $<$.001 & $<$.001 && $<$.001 & $<$.001 & $<$.001 && $<$.001 & $<$.001 & $<$.001 \\
 \multicolumn{2}{c}{DeBERTa} & $<$.001 & $<$.001 & $<$.001 && $<$.001 & $<$.001 & $<$.001 && $<$.001 & $<$.001 & $<$.001 \\
 \multicolumn{2}{c}{BigBird} & $<$.001 & $<$.001 & $<$.001 && $<$.001 & $<$.001 & $<$.001 && $<$.001 & $<$.001 & $<$.001 \\
 \hline
\end{tabular}%
}
\caption{P values corresponding to the `Non-Negative Gain' results in Figure \ref{table:rq3-threshold}, when comparing to random performance of 25\%.}
\end{table}

\subsection*{Exact values for RQ3}

\begin{table}[H]
\centering
\resizebox{\columnwidth}{!}{%
\begin{tabular}{ cccccccccccc }
 \hline
 Fine-tune on $\rightarrow$ & Card &&&& Coin &&&& Dice &&\\
 Test on $\rightarrow$ & Card & Coin & Dice && Card & Coin & Dice && Card & Coin & Dice \\ 
 $\downarrow$ Model& \multicolumn{11}{c}{Train Set(Acc)/Test Set(Acc)} \\
 \hline
 BERT & 100/100 & 75/76 & 75/77 && 75/82 & 100/100 & 50/50 && 85/83 & 74/69 & 100/99\\
 RoBERTa & 100/100 & 75/75 & 75/75 && 75/72 & 100/97 & 50/42 && 75/73 & 75/73 & 100/96\\
 DeBERTa & 100/98 & 96/88 & 99/93 && 100/95 & 100/97 & 75/67 && 72/58 & 100/83 & 85/85\\
 BigBird & 99/90 & 54/56 & 71/67 && 81/59 & 100/100 & 100/97 && 65/63 & 51/50 & 97/91\\
 \hline
\end{tabular}%
}
\caption{Exact quantitative results corresponding to the visualization in Figure \ref{table:rq2-result-fig}}
\label{table:rq2-result-ft}
\end{table}

\end{document}